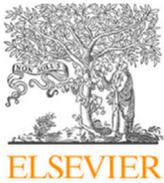
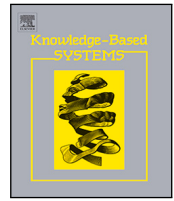

# Artificial Protozoa Optimizer (APO): A novel bio-inspired metaheuristic algorithm for engineering optimization

Xiaopeng Wang [a], Václav Snášel [a,*], Seyedali Mirjalili [b], Jeng-Shyang Pan [c], Lingping Kong [a], Hisham A. Shehadeh [d]

[a] *Faculty of Electrical Engineering and Computer Science, VŠB-Technical University of Ostrava, Ostrava, 70800, Czech Republic*
[b] *Centre for Artificial Intelligence Research and Optimisation, Torrens University Australia, Brisbane, 4006, Australia*
[c] *School of Artificial Intelligence, Nanjing University of Information Science and Technology, Nanjing, 210044, China*
[d] *College of Computer Science and Informatics, Amman Arab University, Amman, 11953, Jordan*

A B S T R A C T

This study proposes a novel artificial protozoa optimizer (APO) that is inspired by protozoa in nature. The APO mimics the survival mechanisms of protozoa by simulating their foraging, dormancy, and reproductive behaviors. The APO was mathematically modeled and implemented to perform the optimization processes of metaheuristic algorithms. The performance of the APO was verified via experimental simulations and compared with 32 state-of-the-art algorithms. Wilcoxon signed-rank test was performed for pairwise comparisons of the proposed APO with the state-of-the-art algorithms, and Friedman test was used for multiple comparisons. First, the APO was tested using 12 functions of the 2022 IEEE Congress on Evolutionary Computation benchmark. Considering practicality, the proposed APO was used to solve five popular engineering design problems in a continuous space with constraints. Moreover, the APO was applied to solve a multilevel image segmentation task in a discrete space with constraints. The experiments confirmed that the APO could provide highly competitive results for optimization problems. The source codes of Artificial Protozoa Optimizer are publicly available at https://seyedalimirjalili.com/projects and https://ww2.mathworks.cn/matlabcentral/fileexchange/162656-artificial-protozoa-optimizer.

## 1. Introduction

The optimization problem exists in many fields such as signal processing, image processing, pattern recognition, mechanical design, production scheduling, and automatic control [1–3]. Optimization refers to the search for optimal solution(s) in a decision space. In general, it maximizes or minimizes one or more objectives subject to certain constraints. In practice, minimizing cost, risk, and time, or maximizing efficiency, profit, and quality are common optimization problems. The properties of the optimization problems can be classified as follows [4,5]. Depending on the type of decision space, they can be classified into two categories: continuous and discrete optimization problems. Considering the constraints, they can be divided into unconstrained and constrained optimizations. In terms of the number of optima, they can be classified as unimodal and multimodal optimization problems. Moreover, the classification distinguishes between single-, multi-, and many-objective optimizations according to the number of objectives. Based on the number of decision variables, they can be divided into low, medium, high dimensions, and large-scale optimizations. In terms of the computational evaluation costs, they can be divided into inexpensive and expensive optimizations. Furthermore, as the environment of optimization problems changes over time (such as changes in objective functions, decision variables, or constraints), optimization problems can also be divided into static and dynamic optimization problems. With the rapid economic development in recent decades, the complexity of solving problems has increased in the scientific and industrial fields, and the need for optimization technologies has become more significant than ever. The mathematical optimization technique is an exact method and used to be the only tool for solving optimization problems. However, this method has certain limitations. The first is premature convergence, which refers to local optima entrapment. The second limitation is that certain algorithms, such as gradient-based algorithms, require derivation and convexity restrictions. However, most problems solved in practical applications are multimodal, nonlinear, and nondifferentiable. Moreover, high computational costs are exhibited when mathematical optimization algorithms are used to solve non-deterministic polynomial-time (NP)-hard problems, where NP-hard implies "at least as hard as the hardest problem in NP". These problems render the mathematical optimization technique ineffective for solving practical problems. Another increasingly popular technique is the metaheuristic algorithm. "Meta" and "heuristic" originate from Greek words. "Meta" implies "upper-level methodology" and "heuristic" means "to discover", "to find", or "to know". A metaheuristic algorithm can be defined as an upper-level general methodology for solving optimization problems [6]. Unlike the exact method, the metaheuristic algorithm is designed to obtain acceptable solutions using an approximate method. The advantages of metaheuristic algorithms are as follows. First, the black-box concept considers only the inputs and outputs of optimization. Second, they can solve large problems, perform faster calculations, and obtain robust results. Moreover, their frameworks and concepts are simple, easy to implement, and flexible. Therefore, metaheuristic algorithms are particularly suitable for optimization problems that are challenging for conventional methods.

* Corresponding author.
 *E-mail address:* vaclav.snasel@vsb.cz (V. Snášel).

Metaheuristic algorithms are generally classified as S-metaheuristics (single solution) and P-metaheuristics (population solutions) [7]. S-metaheuristics begin with a single random solution and iteratively improves it in the decision space. These algorithms exhibit low computational cost and require fewer function evaluations, but suffer from premature convergence, leading to trapping in the local optima. Well-known algorithms in S-metaheuristics include simulated annealing [8], tabu search [9], hill climbing [10], iterated local search [11], and variable neighborhood search [12]. Unlike S-metaheuristic techniques, P-metaheuristic algorithms iteratively enhance a population solution. The process begins with the initialization of a population. Subsequently, a new population is generated based on the current population. Ultimately, the integration of these two populations leads to the subsequent generation of population through diverse selection schemes. The process is terminated, and the discovered optima are output when a predefined condition is satisfied (convergence accuracy, maximum iterations, or maximum function evaluations). In contrast to S-metaheuristics, in terms of performance, P-metaheuristics have a higher ability to avoid local optima. Furthermore, candidate solutions share information during optimization, which helps in solving different difficulties in the decision space. However, high computational cost and additional functional evaluations are two disadvantages of P-metaheuristics.

Biology, physical phenomena, and human behavior usually motivate P-metaheuristics. Several popular P-metaheuristic algorithms are primarily classified into evolution-, swarm-, physics-, and human-based algorithms. Evolution-based algorithms are inspired by biological evolution. The most popular algorithm is the genetic algorithm (GA) [13] proposed by Holland in 1975. It was inspired by Darwin's theory and introduced selection, crossover, and mutation operations to guide the optimization process. In 1995, Storn and Price proposed differential evolution (DE) [14], which was derived from the genetic annealing algorithm [15], where mutation, crossover, and selection operations were integrated for optimization. In 2013, Pinar presented a back-tracking search algorithm (BSA) [16]. The strategy of the BSA is to use a memory population (previous experience) to generate a trial population, including a new crossover and mutation. Rao proposed a simple algorithm called Jaya (which means victory in Sanskrit) [17] in 2016. Jaya updates the population in a significantly simple manner, that is, close to the best solution and far from the worst solution. Swarm-based algorithms draw inspiration from the social behaviors observed in biological swarms. In 1995, Kennedy and Eberhart pioneered the widely recognized particle swarm optimization (PSO) [18]. In PSO, which is inspired by flocks of birds or schools of fish, the significance of both the personal best (called "pbest") and global best (called "gbest") is crucial in directing the population. Mirjalili proposed the grey wolf optimizer (GWO) [19] in 2014. The GWO was inspired by from the leadership and hunting strategies observed in grey wolves. This algorithm introduced four wolf hierarchies: alpha, beta, delta, and omega, along with three hunting behaviors: searching, encircling, and attacking. In 2016, he also proposed a whale optimization algorithm (WOA) [20]. The social behavior and spiral bubble-net feeding strategy of humpback whales were simulated. Song et al. advocated the phasmatodea population evolution (PPE) algorithm [21] in 2021. The authors considered the competitive behavior, population growth, convergent evolution, and path dependence in stick insects. Evolutionary trends guide the population updates to create the next generation. Physics-based algorithms are motivated by the principles of physics, chemistry, and mathematics. A well-known algorithm is the gravitational search algorithm (GSA) [22]. In 2009, Rashedi conceptualized candidate solutions as an assembly of masses, where the interaction among these solutions is governed by Newtonian gravity and motion laws. This interaction dictates the creation of subsequent generations. Mirjalili developed multi-verse optimizer (MVO) [23] in 2015. The MVO incorporates concepts from cosmology by introducing white holes, black holes, and wormholes. In this framework, individual candidate solutions are regarded as distinct universes, with each variable representing an object within that universe. The movement of the population is directed by the repulsion, attraction, and transmission interactions among the holes. In 2016, he also proposed the sine cosine algorithm (SCA) [24]. The SCA randomly initializes multiple candidate solutions and constructs mathematical models based on sine and cosine functions to regulate their movement either outward or toward the best solution. In 2021, Abualigah introduced the arithmetic optimization algorithm (AOA) [25], which implements optimization by utilizing addition, subtraction, multiplication, and division operations. Human-inspired algorithms are influenced by human social behavior. A notable instance is teaching–learning-based optimization (TLBO) [26], which was introduced by Rao in 2011. In the TLBO, solutions are viewed as a collective of learners, encompassing both teachers and students. The algorithm employs "Teacher Phase" and "Learner Phase" operations to boost the population. During the "Teacher Phase", students learn from the teacher, whereas the "Learner Phase" involves learning through interactions among students. In 2019, Zhao described supply—demand-based optimization (SDO) [27], a proposal inspired by the economic principles of supply and demand. Mathematical modeling is based on the interplay between consumer demand and producer supply. Kaveh presented an art-motivated stochastic paint optimizer (SPO) [28] in 2020. The SPO treats the decision space as a canvas, where the population search is driven by four basic color combinations to uncover the most beautiful combination pattern. Student psychology-based optimization (SPBO) [29] was proposed by Das in 2020. This model is based on the strategy that students work harder to be at the top of their class. This paper outlines four student categories within a class: best, good, and average students, and students trying to improve randomly. The efforts of the students will result in an increase in their overall grade. Numerous other P-metaheuristics are worth acknowledging, including ant colony optimization [30], gaining-sharing knowledge-based algorithm [31], dwarf mongoose optimization [32], nutcracker optimization algorithm [33], snake optimizer (SO) [34], and harris hawks optimizer (HHO) [35], quadratic interpolation optimization (QIO) [36], and Fick's law optimization [37].

Metaheuristic algorithms are inspired by various natural phenomena and operate in different ways. However, the two common concepts are exploration (diversification) and exploitation (intensification). In [38], Eiben and Schippers state that "exploration and exploitation are the two cornerstones of problem solving by search". Exploration pertains to the capability of generating diverse solutions to explore the global space, whereas exploitation emphasizes the ability to conduct a local search around promising solutions. Exploration and exploitation are conflicting components, and the preference for exploration inevitably leads to the degradation of exploitation, and vice versa. Despite its significance, a specific guideline has not been proposed for achieving this balance. Consequently, determining the right balance between exploration and exploitation remains a challenging task. "No Free Lunch" [39] theory asserts that no heuristic can consistently outperform others across all possible problems. Even with a purely random search, the average performance will be comparable on all problems for different heuristics. This implies that an algorithm may be effective for specific problems; however, its performance may degrade when applied to different problems. Consequently, this theorem has opened up the field and sparked significant interest among researchers proposing new algorithms. The primary research and contributions of this paper are as follows.

- A new bio-inspired artificial protozoa optimizer (APO) is designed to model the survival behavior of protozoa.
- The APO algorithm mimics foraging, dormancy, and reproduction. Autotrophic foraging and dormancy contribute to exploration, however heterotrophic foraging and reproduction contribute to exploitation.
- The APO is implemented and evaluated under the CEC2022 benchmark. The experimental results verified that the APO is superior to 32 state-of-the-art algorithms.
- The effectiveness of APO is tested by challenging real-world problems, including five engineering designs and a multilevel image segmentation task.

The remainder of this paper is structured as follows. In Section 2, the proposed APO algorithm is introduced. Section 3 presents simulations under the CEC2022 benchmark. Section 4 highlights certain practical applications, including five engineering designs and a multilevel image segmentation task. Finally, Section 5 summarizes the study and proposes potential research directions.

## 2. Artificial protozoa optimizer

In this section, we introduce the APO algorithm, beginning with an elucidation of its inspiration. Subsequently, we propose mathematical models for simulating protozoa. Finally, a comprehensive analysis of the proposed algorithm is presented.

### 2.1. Inspiration

The investigation of various biological phenomena demonstrates the distinct advantages of microorganisms. Bacteria, algae, and protozoa within microorganisms perform analogous functions to that of the organs found in higher plants and animals, which are accomplished through specialized structures known as "organelles". These microorganisms display fundamental life characteristics including metabolism, reproduction, genetic continuity, variability, and adaptation to environmental stimuli. Microorganisms are typically utilized more efficiently than higher organisms because of their simpler organization and low complexity [40].

In this paper, the mentioned protozoa refer to the representative euglena within the flagellates. The name "euglena" is derived from a Greek word meaning "eyeball organism". It is a single-celled protozoan or, more precisely, an "algal flagellate". More than 250 species of euglena have been identified. Euglena species primarily reside in freshwater environments and often form prolific blooms in ponds and ditches. The abundant growth of this species imparts a green or red tint to the water surface. Euglena cells range in size from 15 to 500 μm, and their shapes vary from nearly spherical to nearly cylindrical. Most euglena species are green because of the presence of chlorophyll a and b in their chloroplasts [41,42]. Fig. 1 shows a spindle-shaped euglena and its prominent organelles containing flagellum, photoreceptor, nucleus, mitochondria, chloroplasts, stigma (eyespot), contractile vacuole, Golgi apparatus, endoplasmic reticulum, stored carbohydrate, and pellicles [43]. Euglena has two flagella that are rooted in its basal body. One short flagellum does not protrude from the cell, while the other is long enough to protrude from the "gullet" or pharynx. The euglena performs spiral motions by "beating" the long flagellum.

Notably, euglena exhibits both plant- and animal-like characteristics. This implies that it can get the nutrition to survive by functioning as an autotrophic (mineral) or a heterotrophic (organic substance) organism. When the euglena is under unfavorable environmental conditions, it forms a protective wall around itself such as a cyst and becomes dormant until the surrounding environment improves. For offspring production, euglena reproduces through asexual reproduction by the process of binary fission, that is, it splits into two through longitudinal cell division. Euglena can be used for environmental biomediation and exhibits significant potential for use in biomedicine. First, euglena can be used as an indicator of environmental health and assess water quality. It is highly resistant to heavy metals and can be used to restore water contaminated by radioactive elements. Moreover, euglena contains a toxin called euglenophycin, which exhibits anticancer activity. It is also a source of single-cell proteins and can be used in a variety of drugs, such as anti-hypertensive, anti-gout, and anti-liver cancer drugs. Finally, euglena demonstrates potential for use in commercial products such as dietary supplements, food preservatives, and cosmetics [44]. The foraging, dormancy, and reproductive behaviors of euglena are further described based on mathematical models in the next subsection.

- Foraging

Euglena absorbs the nutrients essential for its survival through both autotrophic and heterotrophic mechanisms. Euglena performs photosynthesis similar to plants, using chloroplasts to produce carbohydrates for energy. Phototaxis is a behavioral response of the euglena to light stimuli, which enables it to move toward or away from light. The perception of light is considered to be the basis of phototaxis. Euglena possesses an organelle known as the eyespot, which consists of carotenoid pigment granules. The eyespot is itself not sensitive to light and functions as a filter for sunlight, permitting only specific wavelengths of light to reach the photoreceptor at the base of the flagellum. At low irradiance, the euglena swims spirally toward the light source. However, the euglena moves away from the light source under high irradiance to prevent the destruction of pigments and chloroplasts. For euglena, the threshold for switching from positive to negative phototaxis has been investigated to be between 10 and 100 W/m$^2$. This indicates that the euglena can self-orient to locate suitable light conditions for photosynthesis. Thus, euglena exhibits positive phototaxis under low light and negative phototaxis under intense light. This causes it to congregate in suitable habitats for survival. In the dark, euglena exhibits animal-like behavior by absorbing organic matter through phagocytosis, including substances such as carbohydrates, peptone, acetate, and beef extract. Euglena differs from other plant cells because it lacks a rigid cellulose wall. Instead, it features a flexible pellicle capable of free deformation. Through osmotrophy, euglena can directly absorb nutrients from the surrounding environment [45,46].

- Dormancy

Organisms survive in various ways under different stresses, such as environmental pollution, temperature changes, or a lack of food. Some animals migrate to reduce environmental stress, whereas others avoid adverse situations by changing their biological behavior. In response to unfavorable conditions, euglena can spray mucus subcutaneously, forming a protective mucilaginous capsule called cyst. During this period, the euglena enters a state of suspended animation (dormancy) until the surrounding environmental conditions become more conducive. The original cell morphs into ovoid, elliptical, and spherical shapes, and its flagella disappear. Euglena transitions from a unicellular motile organism to a non-motile mucus colony. In this suspended animation, the metabolism of the euglena slows down, resulting in energy conservation and reduced reliance on the external environment. Dormancy is an adaptive capability of euglena to effectively respond to environmental stress [47].

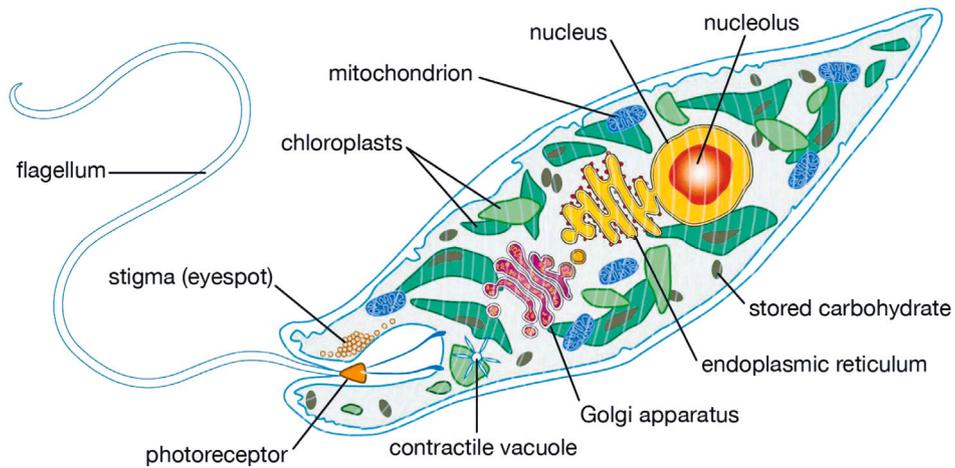

**Fig. 1.** Euglena cell showing the main organelles.

**Table 1**
Lowercase indicates a scalar parameter and uppercase indicates a vector.

| | |
|---|---|
| $ps$ | Population size. |
| $dim$ | Number of decision variables. |
| $di$ | Dimension index, $di \in \{1, 2, \ldots, dim\}$. |
| $np$ | Number of neighbor pairs. |
| $f$ | Foraging factor. |
| $w_a$ | Weight factor in autotroph. |
| $w_h$ | Weight factor in heterotroph. |
| $pf$ | Proportion fraction of dormancy and reproduction. |
| $p_{ah}$ | Probability of autotrophic and heterotrophic behavior. |
| $p_{dr}$ | Probability of dormancy and reproduction. |
| $rand$ | Random number in the range [0,1]. |
| $iter$ | Current iteration. |
| $iter_{max}$ | Maximum iteration. |
| $X_i$ | $i$th protozoan. |
| $X_{min}$ | Lower-bound vector. |
| $X_{max}$ | Upper-bound vector. |
| $M_f$ | Mapping vector in foraging. |
| $M_r$ | Mapping vector in reproduction. |
| $Dr_{index}$ | Index vector in dormancy and reproduction. |
| $Rand$ | Random vector with elements in [0,1]. |
| $\lceil \cdot \rceil$ | Ceiling function. |
| $\lfloor \cdot \rfloor$ | Flooring function. |
| $f(\cdot)$ | Fitness function. |
| $sort(\cdot)$ | Ranking is based on fitness values from smallest to largest. |
| $randperm(n, l)$ | Return a row vector containing $l$ unique integers that are randomly selected between 1 and $n$. |
| $\odot$ | Hadamard product. |

• Reproduction

Euglena exhibits a simple form of asexual reproduction known as binary fission. During this process, the euglena splits into two identical individuals. The optimal temperature range for euglena reproduction is between 20–35 °C. Binary fission commences with mitosis of the euglena nucleus and involves the replication of the flagellum, esophagus, and stigma. The subsequent division occurs along the longitudinal axis of the cell. Initially, a cleft is formed at the front part, and then the V-shaped bifurcation gradually progresses toward the posterior until the two halves are completely separated [48].

*2.2. Mathematical models*

In this subsection, the algorithm to solve the minimization problem is introduced. Table 1 provides a list of notations and nomenclature that are used. The representation of a solution set is essential for metaheuristic algorithms. In our proposed algorithm, the protozoa represent the solution set, and each protozoan has its position consisting of $dim$ variables.

*2.2.1. Foraging*

For foraging behavior, we considered the internal and external factors of the protozoa. Internal factors are regarded as the foraging characteristics of the protozoa, whereas external factors are considered as environmental influences such as species collisions and competitive behaviors.

• Autotrophic mode

As mentioned previously, a protozoan can produce carbohydrates through chloroplasts to supply nutrition. If the protozoan is exposed to strong light intensity, it will move away from its position and move toward a location with a lower light intensity. The converse is true when it is in a location with low light intensity. Assuming that the light intensity around the $j$th protozoan is suitable for photosynthesis, the protozoan will move to the location of the $j$th protozoan. For the autotrophic mode, we provide the following mathematical model:

$$X_i^{new} = X_i + f \cdot (X_j - X_i + \frac{1}{np} \cdot \sum_{k=1}^{np} w_a \cdot (X_{k-} - X_{k+})) \odot M_f \quad (1)$$

$$X_i = [x_i^1, x_i^2, \ldots, x_i^{dim}], \qquad X_i = sort(X_i) \quad (2)$$

$$f = rand \cdot (1 + \cos(\frac{iter}{iter_{max}} \cdot \pi)) \tag{3}$$

$$np_{max} = \lfloor \frac{ps-1}{2} \rfloor \tag{4}$$

$$w_a = e^{-\left|\frac{f(X_{k-})}{f(X_{k+})+eps}\right|} \tag{5}$$

$$M_f[di] = \begin{cases} 1, & \text{if } di \text{ is in } randperm(dim, \lceil dim \cdot \frac{i}{ps} \rceil) \\ 0, & \text{otherwise} \end{cases} \tag{6}$$

where $X_i^{new}$ and $X_i$ denote the updated position, and original position of the $i$th protozoan, respectively. $X_j$ is the randomly selected $j$th protozoan. $X_{k-}$ denotes a randomly selected protozoan in the $k$th paired neighbor whose rank index is less than $i$. Specifically, if $X_i$ is $X_1$, $X_{k-}$ is also set as $X_1$. $X_{k+}$ denotes a randomly selected protozoan in the $k$th paired neighbor, and its rank index is greater than $i$. Particularly, if $X_i$ is $X_{ps}$, $X_{k+}$ is also set to $X_{ps}$, where $ps$ is the population size. $f$ represents a foraging factor and $rand$ denotes a random number in the [0,1] interval from the uniform distribution. $iter$ and $iter_{max}$ denote the current and maximum iterations, respectively. $np$ indicates the number of neighbor pairs among the external factors and $np_{max}$ is the maximum value of $np$. $w_a$ is a weight factor in the autotrophic mode and $eps$ (2.2204e−16) is a significantly small number. $\odot$ denotes the Hadamard product. $M_f$ is a mapping vector for foraging with a size of $(1 \times dim)$, where each element is 0 or 1. $di$ denotes the dimensional index $di \in \{1, 2, \ldots, dim\}$.

- Heterotrophic mode

In the dark, a protozoan can obtain nutrients by absorbing organic matter from its surroundings. Assuming that $X_{near}$ is a nearby food-rich location, the protozoan moves toward it. For the heterotrophic mode, we propose the following mathematical model:

$$X_i^{new} = X_i + f \cdot (X_{near} - X_i + \frac{1}{np} \cdot \sum_{k=1}^{np} w_h \cdot (X_{i-k} - X_{i+k})) \odot M_f \tag{7}$$

$$X_{near} = (1 \pm Rand \cdot (1 - \frac{iter}{iter_{max}})) \odot X_i \tag{8}$$

$$w_h = e^{-\left|\frac{f(X_{i-k})}{f(X_{i+k})+eps}\right|} \tag{9}$$

$$Rand = [rand_1, rand_2, \ldots, rand_{dim}] \tag{10}$$

where $X_{near}$ is a nearby location, and "±" implies that $X_{near}$ can be in different directions from the $i$th protozoan. $X_{i-k}$ denotes the $(i-k)$th protozoan selected from the $k$th paired neighbor, and its rank index is $i-k$. Specifically, if $X_i$ is $X_1$, $X_{i-k}$ is also set to $X_1$. $X_{i+k}$ denotes the $(i+k)$th protozoan selected from the $k$th paired neighbor, and its rank index is $i+k$. Particularly, if $X_i$ is $X_{ps}$, $X_{i+k}$ is also set to $X_{ps}$. $w_h$ is the weight factor in the heterotrophic mode. $Rand$ is a random vector with elements in the [0,1] interval.

### 2.2.2. Dormancy

During environmental stress, a protozoan may adopt dormant behavior as a survival strategy to endure unfavorable conditions. When the protozoan is dormant, it is replaced by a newly generated protozoan to maintain a constant population. The mathematical model for dormancy is as follows:

$$X_i^{new} = X_{min} + Rand \odot (X_{max} - X_{min}) \tag{11}$$

$$X_{min} = [lb_1, lb_2, \ldots, lb_{dim}], \quad X_{max} = [ub_1, ub_2, \ldots, ub_{dim}] \tag{12}$$

where $X_{min}$ and $X_{max}$ represent the lower- and upper-bound vectors, respectively. $lb_{di}$ and $ub_{di}$ indicate the lower and upper bounds of the $di$th variable, respectively.

### 2.2.3. Reproduction

At an appropriate age and health, protozoa undergo asexual reproduction, which is known as binary fission. Theoretically, this reproduction causes the protozoan to split into two identical daughters. We simulated this behavior by generating a duplicate protozoan and considering a perturbation. The mathematical model for reproduction is as follows:

$$X_i^{new} = X_i \pm rand \cdot (X_{min} + Rand \odot (X_{max} - X_{min})) \odot M_r \tag{13}$$

$$M_r[di] = \begin{cases} 1, & \text{if } di \text{ is in } randperm(dim, \lceil dim \cdot rand \rceil) \\ 0, & \text{otherwise} \end{cases} \tag{14}$$

where "±" implies the perturbation can be forward and reverse. $M_r$ is a mapping vector in the reproduction process, whose size is $(1 \times dim)$, and each element is 0 or 1.

### 2.2.4. Algorithm

The details of the APO are presented below. To integrate all the mathematical models, the parameters involved are as follows:

$$pf = pf_{max} \cdot rand \tag{15}$$

$$p_{ah} = \frac{1}{2} \cdot (1 + \cos(\frac{iter}{iter_{max}} \cdot \pi)) \tag{16}$$

$$p_{dr} = \frac{1}{2} \cdot (1 + \cos((1 - \frac{i}{ps}) \cdot \pi)) \tag{17}$$

where $pf$ is a proportion fraction of dormancy and reproduction in the protozoa population and $pf_{max}$ is the maximum value of $pf$. $p_{ah}$ indicates the probabilities of autotrophic and heterotrophic behaviors, and $p_{dr}$ indicates the probabilities of dormancy and reproduction.

Note that the proposed APO has only two special parameters: $np$ (number of neighbor pairs) and $pf_{max}$ (maximum proportion fraction). The framework of the proposed algorithm is shown in Fig. 2, and the corresponding pseudo code is outlined in Algorithm 1.

### 2.3. Algorithm analysis

In this subsection, the proposed APO algorithm is analyzed by considering the algorithm design, search operators, exploration and exploitation, and computational complexity.

#### 2.3.1. A discussion on APO algorithm design

The design of metaheuristic algorithms is inspired by nature. In the last 20 years, hundreds of new algorithms have been proposed by researchers to enable more successful search performance. Fig. 3 illustrates the general process and frequently used methods of metaheuristic algorithms. Metaheuristic algorithms can be divided into two stages to be analyzed. The first stage is the initialization of the population, with random sampling and Latin hypercube sampling as the main manners. Once the initial population is generated, each candidate solution's fitness value is evaluated. The second phase is an iterative search process for the population, associated with three steps. The first step is the guidance mechanism, this step solves how to select candidate solutions from the population as reference positions in the search space. Each reference position can be a candidate solution or a combination of multiple candidate solutions. The random method employs a non-deterministic approach to select candidate solutions, and this random selection contributes to the diversity of the population search. The greedy method selects candidate solutions with better fitness values to guide the search of the population. This method focuses more on convergence and searches around promising candidate solutions. The probabilistic method takes into account the characteristics of both the random and greedy selection, and a candidate solution with a better fitness value is more likely to be selected. Probabilistic methods commonly used are the roulette wheel and tournament methods. The last one is the ordinal method. This selection focuses on the relative order of indexes in the population, sequentially selecting each candidate solution as a reference position. The second step is the search operators, which model the unique behavior of natural populations. The algorithms differ from each other, but these models serve the exploration and exploitation tasks. Exploration maintains the diversity of candidate solutions and prevents premature convergence to a local optimum. However, exploitation emphasizes the search around the reference positions to find the optimum to accelerate convergence. The update mechanism is the third step. This is to define the new population, selecting candidate solutions for the next generation to search. The first one is the direct method, no pre-processing process is performed and the new candidate solutions are directly selected to the next generation. The second method, fitness value-based selection, is the most frequently used. This method is based on Darwin's theory of evolution, where the principle of "survival of the fittest" determines that better candidate solutions are selected to the next generation. The meta-heuristic algorithms will provide the optimal solutions found after the second iteration stage. Recently, some researchers investigated the guidance mechanism and update mechanism in metaheuristic algorithms, and proposed several promising methods: fitness-distance balance (FDB) [49], adaptive fitness-distance balance (AFDB) [50], fitness-distance-constraint (FDC) [51], dynamic fitness-distance balance (DFDB) [52], and natural survivor method (NSM) [53]. These new methods can be introduced into the design of metaheuristic algorithms.

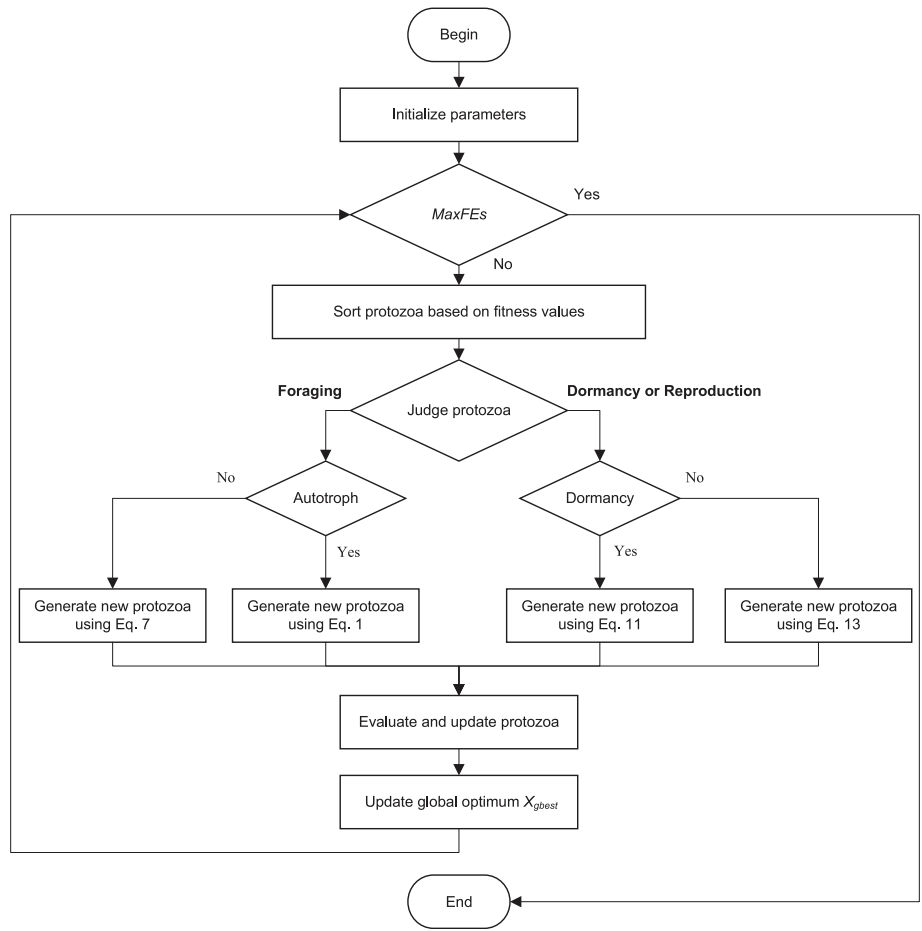

**Fig. 2.** The framework of artificial protozoa optimizer.

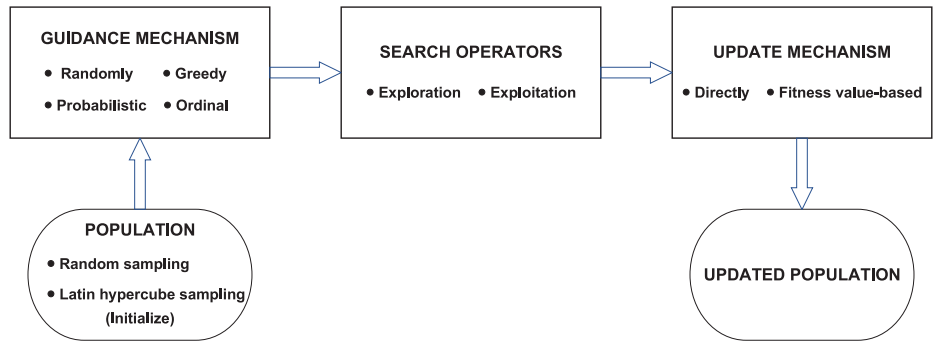

**Fig. 3.** The general process of metaheuristic algorithms.

For the APO algorithm proposed in this paper, the initialization is based on traditional random sampling; the guidance mechanism is an ordinal selection method; the update mechanism uses the fitness value-based method; and for the important exploration and exploitation phases, the related search operators are next described in detail.

**Algorithm 1** Pseudo code of the proposed artificial protozoa optimizer

**Input:** Initialize parameters $ps$, $dim$, $np$, $pf_{max}$, and $MaxFEs$ (maximum function evaluations).
**Output:** The global optima $X_{gbest}$ and $f(X_{gbest})$.
1: **while** $FEs < MaxFEs$ **do**
2:    $sort(X_i)$,    $i = 1, 2, \ldots, ps$;
3:    $pf = pf_{max} \cdot rand$;    // proportion fraction
4:    $Dr_{index} = randperm(ps, \lceil ps \cdot pf \rceil)$;    // index vector of dormancy and reproduction
5:    **for** $i = 1 : ps$ **do**
6:       **if** $i$ is in $Dr_{index}$ **then**
7:          **if** $p_{dr} > rand$ **then**
8:             Calculate $X_i^{new}$ using Eq. (11);    // dormancy.
9:          **else**
10:             $M_r = zeros(1, dim)$;
11:             $M_r[1, randperm(dim, \lceil dim \cdot rand \rceil)] = 1$;
12:             Calculate $X_i^{new}$ using Eq. (13);    // reproduction
13:          **end if**
14:       **else**
15:          $M_f = zeros(1, dim)$;
16:          $M_f[1, randperm(dim, \lceil dim \cdot \frac{i}{ps} \rceil)] = 1$;
17:          **if** $p_{ah} > rand$ **then**
18:             Calculate $X_i^{new}$ using Eq. (1);    // foraging in an autotroph
19:          **else**
20:             Calculate $X_i^{new}$ using Eq. (7);    // foraging in a heterotroph
21:          **end if**
22:       **end if**
23:       **if** $f(X_i^{new}) < f(X_i)$ **then**
24:          $X_i \leftarrow X_i^{new}$;
25:       **else**
26:          $X_i \leftarrow X_i$;
27:       **end if**
28:    **end for**
29:    $X_{gbest} = opt\{X_i\}$;
30:    $FEs \leftarrow FEs + ps$;
31: **end while**

*2.3.2. Search operators*

For a clear understanding of the mathematical models, Fig. 4 shows the search operators in two dimensions ($np = 1$). Note that this approach can be extended to higher dimensions.

- Autotroph (exploration)

In Fig. 4(a), $f$ is the foraging factor. For $f > 1$, $f = 1$, or $f < 1$, the updated $X_i^{new}$ is represented by blue, red, or green points, respectively. The following analysis is based on the case where $f = 1$. $X_i$ denotes the position of the $i$th protozoan and $X_j$ denotes a randomly selected $j$th protozoan. A rank index of $j < i$ implies that the $i$th protozoan is at a lower light intensity and exhibits phototaxis; if the rank index $j > i$, then the $i$th protozoan is at a higher light intensity and exhibits photophobia. Conversely, if the rank index $j = i$, then the $i$th protozoan is under a suitable light intensity for photosynthesis. The vector $\overline{X_i X_j}$ simulates the internal factor of the $i$th protozoan in the autotrophic mode. $X_{k-}$ and $X_{k+}$ denote the randomly selected protozoa in the $k$th paired neighbor ($np$ is set to 1, implying that only one pair of neighbors is considered). The rank index of $X_{k-}$ is less than $i$ and the rank index of $X_{k+}$ is greater than $i$. Particularly, if $X_i$ is $X_1$, $X_{k-}$ is also set to $X_1$, and if $X_i$ is $X_{ps}$, $X_{k+}$ is also set to $X_{ps}$. $w_a$ is the weight factor in the autotrophic mode. The vector $\overline{X_j X_i^{new}}$ simulates the external factors of the $i$th protozoan. $M_f$ is a mapping vector with elements 0 or 1 in foraging. $M_f = [1, 1]$ implies that both dimensions of the $i$th protozoan are updated. $M_f = [1, 0]$ implies that only the first dimension of the $i$th protozoan is updated. $M_f = [0, 1]$ implies that only the second dimension is updated. Eventually, one of the three red points $X_i^{new}$ is selected based on $M_f$ to update the $i$th protozoan.

- Heterotroph (exploitation)

As shown in Fig. 4(b), $X_i$ denotes the position of the $i$th protozoan and $X_{near}$ denotes a location near that of the $i$th protozoan. The vector $\overline{X_i X_{near}}$ simulates the internal factor of the $i$th protozoan in the heterotrophic mode. $X_{i-k}$ and $X_{i+k}$ denote the selected protozoa in the $k$th paired neighbor ($np$ is set to 1, which implies that only one pair of neighbors is considered). The rank index of $X_{i-k}$ is $i-k$, and that of $X_{i+k}$ is $i + k$. Particularly, if $X_i$ is $X_1$, $X_{i-k}$ is set to $X_1$; if $X_i$ is $X_{ps}$, $X_{i+k}$ is also set to $X_{ps}$. $w_h$ is the weight factor in the heterotrophic mode. The vector $\overline{X_{near} X_i^{new}}$ simulates the external factors of the $i$th protozoan. The foraging factor $f$ and mapping vector $M_f$ refer to the autotrophic mode. Finally, one of the three red points of $X_i^{new}$ is selected based on $M_f$ to update the $i$th protozoan.

- Dormancy (exploration)

As shown in Fig. 4(c), $X_i$ denotes the position of the $i$th protozoan. When the $i$th protozoan encounters environmental stress, it becomes dormant to survive. A randomly generated new protozoan will replace it to maintain a constant population. Therefore, the $i$th protozoan, $X_i$ is updated to $X_i^{new}$.

- Reproduction (exploitation)

As shown in Fig. 4(d), $X_i$ denotes the position of the $i$th protozoan. The vector $\overline{X_i X_i^{new}}$ represents a perturbation ($\Delta$) that simulates a reproductive mutation. $M_r$ is a mapping vector with elements 0 or 1 during reproduction. If $M_r = [1, 1]$, both the dimensions of the $i$th protozoan are updated; if $M_r = [1, 0]$, only the first dimension is updated; and if $M_r = [0, 1]$, only the second dimension is updated. Thus, one of the three red points of $X_i^{new}$ is selected based on $M_r$ to update the $i$th protozoan.

*2.3.3. Exploration and exploitation*

The effectiveness of metaheuristic algorithms depends on a balance between exploration and exploitation. Six parameters ($pf$, $p_{ah}$, $p_{dr}$, $f$, $M_f$, and $M_r$) play important roles in the exploration and exploitation.

Fig. 5 illustrates the exploration and exploitation phases of the proposed APO algorithm. The initial state of the protozoa is based on the proportion fraction ($pf$), which categorizes whether the protozoa are involved in foraging (autotroph and heterotroph), dormancy, or reproduction. Autotroph and dormancy focus on large-coverage searches to improve the exploration ability of the algorithm. Conversely, heterotroph and reproduction search for promising areas to improve the exploitation ability. $p_{ah}$ is a probability parameter for autotroph and heterotroph. If $p_{ah} > rand$, then the protozoa are in autotrophic mode. Otherwise, they are in heterotrophic mode. As the number of iterations increases, $p_{ah}$ gradually decreases, shifting the focus from autotroph to heterotroph. $p_{dr}$ is the probability parameter for dormancy and reproduction. If $p_{dr} > rand$, then the protozoa are dormant. Otherwise, they are in reproductive mode. $p_{dr}$ is related to the ranking of the protozoans. The inferior protozoa tend to perform dormancy to enhance their exploration. Conversely, the superior protozoa tend to reproduce to enhance their exploitation. The emphasis on exploration by the parameter $p_{dr}$ extends not only during the initial iteration but also throughout the final iteration, providing significant assistance in mitigating the local optimal stagnation. $f$ is a foraging factor that controls the search distance of the protozoa during foraging. The dynamic characteristics are shown in Fig. 6. As the number of iterations increases, the value of $f$ gradually decreases from 2 to 0. This results in the search distance transitioning from blue to red points, and then to green points in

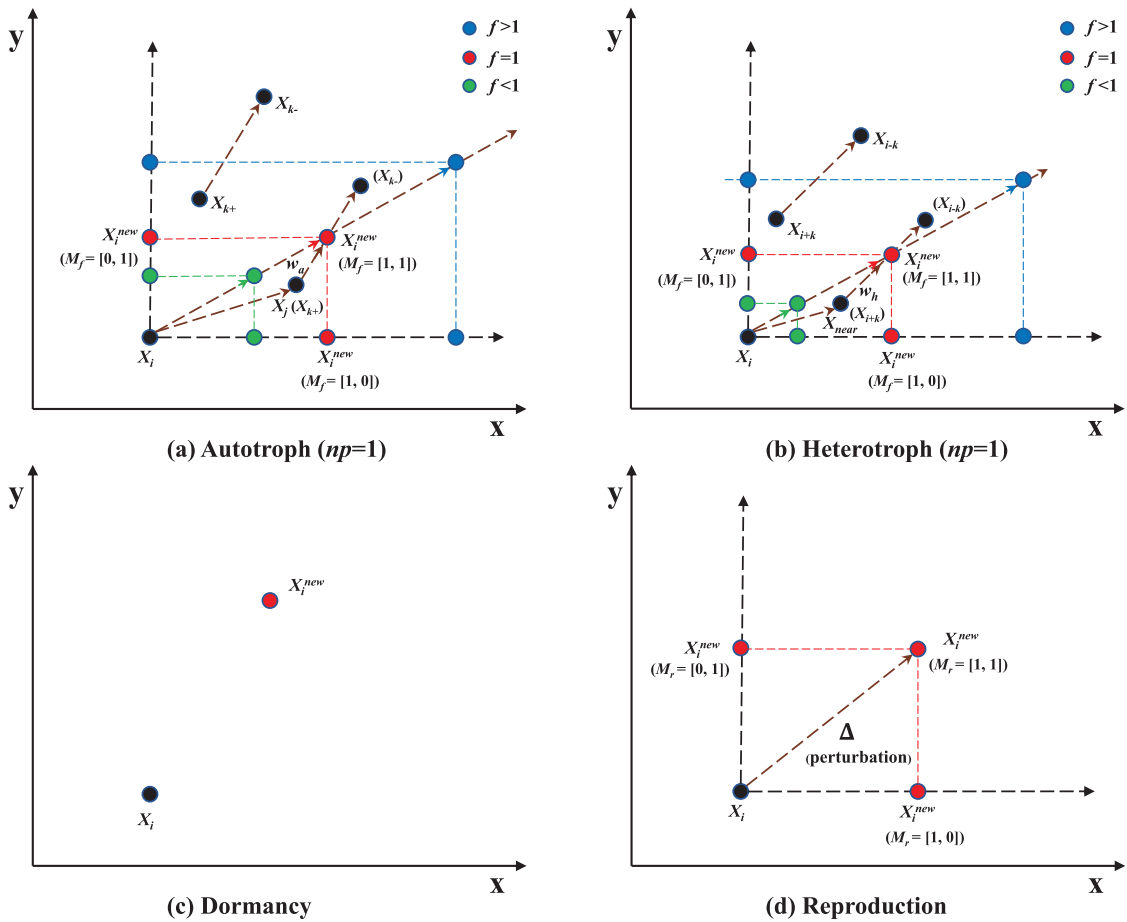

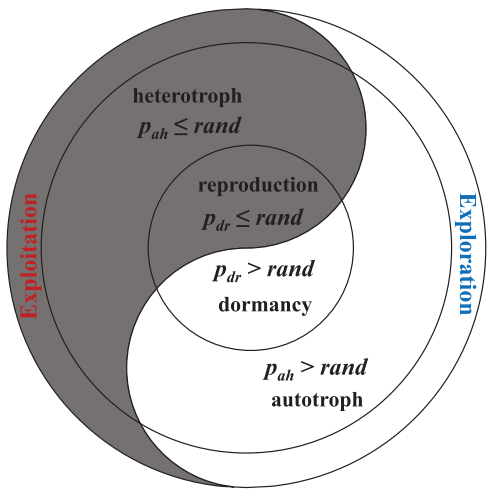

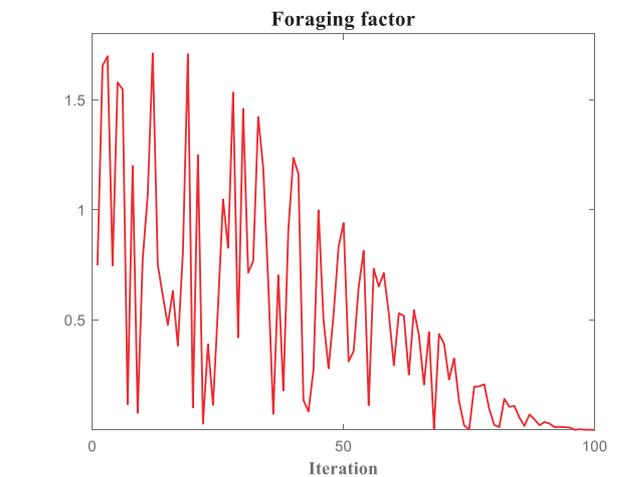

**Fig. 4.** Search operators.

**Fig. 5.** Exploration and exploitation.

**Fig. 6.** Dynamic character of foraging factor during 100 iterations.

Figs. 4(a) and (b). This transition is beneficial for balancing exploration and exploitation. $M_f$ and $M_r$ are the mapping vectors that control the updated dimensions of the protozoa. $M_f$ is related to the ranking of the protozoa. The dimensions of the low-ranked protozoa changed more, whereas the dimensions of the top-ranked protozoa changed less. This causes the inferior protozoa to focus on exploration and the superior protozoa to focus on exploitation. $M_r$ is a random mapping vector that affects the mutated dimensions. It performs an exploitation to focus on searching nearby locations for the protozoa.

### 2.3.4. Computational complexity

The proposed APO algorithm primarily performs sorting, autotroph and heterotroph, dormancy and reproduction, and fitness function evaluation. The computational complexity of sorting is $O(ps \cdot \log(ps))$, autotroph and heterotroph is $O((ps - \lceil ps \cdot pf \rceil) \cdot np \cdot dim)$, dormancy and reproduction is $O(\lceil ps \cdot pf \rceil \cdot dim)$, and fitness function evaluation is $O(ps \cdot f(\cdot))$. Consequently, the total complexity is $O(iter_{max} \cdot (ps \cdot \log(ps) + ((ps - \lceil ps \cdot pf \rceil) \cdot np + \lceil ps \cdot pf \rceil) \cdot dim + ps \cdot f(\cdot)))$. In particular, if $np = 1$, then the complexity of the APO algorithm is $O(iter_{max} \cdot ps \cdot (\log(ps) + dim + f(\cdot)))$.

**Table 2**
Benchmark functions of the 2022 IEEE Congress on Evolutionary Computation (CEC2022).

|  | No. | Functions | $f_{min}$ |
| --- | --- | --- | --- |
| Unimodal function | F1 | Shifted and full Rotated Zakharov function | 300 |
| Multimodal functions | F2 | Shifted and full Rotated Rosenbrock's function | 400 |
|  | F3 | Shifted and full Rotated Expanded Schaffer's f6 function | 600 |
|  | F4 | Shifted and full Rotated Non-Continuous Rastrigin's function | 800 |
|  | F5 | Shifted and full Rotated Levy function | 900 |
| Hybrid functions | F6 | Hybrid function 1 ($N = 3$) | 1800 |
|  | F7 | Hybrid function 2 ($N = 6$) | 2000 |
|  | F8 | Hybrid function 3 ($N = 5$) | 2200 |
| Composition functions | F9 | Composition function 1 ($N = 5$) | 2300 |
|  | F10 | Composition function 2 ($N = 4$) | 2400 |
|  | F11 | Composition function 3 ($N = 5$) | 2600 |
|  | F12 | Composition function 4 ($N = 6$) | 2700 |
|  | Search range: $[-100, 100]^{dim}$ | | |

**Table 3**
Parameter setting of the comparison algorithms.

|  | Algorithm | Parameter settings | Year |
| --- | --- | --- | --- |
| Evolutionary-based algorithm | GA | selection - roulette wheel, crossover - whole arithmetic, mutation - uniform mutation, mutation rate = 0.2 | 1975 |
|  | DE | $f = 0.7, cr = 0.1$ | 1995 |
|  | BSA | $mixrate = 1$ | 2013 |
|  | Jaya | no special parameters | 2016 |
| Swarm-based algorithm | PSO | $v_{min} = -10, v_{max} = 10, w = [0.9, 0.4], c_1, c_2 = 2$ | 1995 |
|  | GWO | $a = [2, 0]$ | 2014 |
|  | WOA | $a = [2, 0]$ | 2016 |
|  | PPE | $a = 1.1, c = 0.2$ | 2021 |
| Physics-based algorithm | GSA | $G_0 = 100, \alpha = 20$ | 2009 |
|  | MVO | $WEP_{min} = 0.2, WEP_{max} = 1, p = 0.6$ | 2015 |
|  | SCA | $a = 2$ | 2016 |
|  | AOA | $\alpha = 5, \mu = 0.5$ | 2021 |
| Human-based algorithm | TLBO | no special parameters, $ps = \frac{1}{2} \cdot$ setting $ps$, because each generation uses $(2 \times ps)$ FEs. | 2011 |
|  | SDO | no special parameters, $ps = \frac{1}{2} \cdot$ setting $ps$, because each generation uses $(2 \times ps)$ FEs. | 2019 |
|  | SPO | no special parameters, $ps = \frac{1}{4} \cdot$ setting $ps$, because each generation uses $(4 \times ps)$ FEs. | 2020 |
|  | SPBO | no special parameters | 2020 |

## 3. Experimental analysis

In this section, the proposed algorithm is evaluated using the CEC2022 benchmark [54]. A detailed definition of the benchmark is provided in Table 2. The experimental simulation was executed on a personal laptop loaded with Windows 11 operating system, equipped with 16 GB RAM, and powered by an Intel(R) Core(TM) i7-8750H CPU @ 2.20 GHz 2.21 GHz. The APO algorithm was coded in MATLAB.

### 3.1. Parameter setting

For the CEC2022 benchmark, the research followed the standards defined at the CEC conference. The number of dimensions of the function ($dim$) was 20, the number of max fitness evaluations was 1,000,000, and each algorithm was executed 30 times. In the proposed APO algorithm, $np$ was set to 1 considering the algorithm complexity. The parameter $pf_{max}$ was systematically tested at $0, 0.1, 0.2, \ldots,$ and 1. The results of the Friedman test method indicated that superior performance was achieved at 0.1. Consequently, $pf_{max}$ was set to 0.1 in this paper. For the compared algorithms, the parameter settings were based on the original references. Table 3 lists the detailed parameters and their corresponding publication years.

### 3.2. The performance of APO at different population scales

This part investigates the impact of different population sizes on the proposed APO algorithm. Despite the same computing resources, variations in performance still exist for the same algorithm. We assessed the APO algorithm under different population sizes by examining three categories: small ($ps = 10, 20$), medium ($ps = 40, 50$), and large ($ps = 80, 100$) populations. Table 4 lists the rankings based on the Friedman test. For example, for the unimodal function (F1), $ps$ 10, $ps$ 20, $ps$ 40, $ps$ 50, $ps$ 80, and $ps$ 100 are ranked sixth, fifth, fourth, third, second, and first, respectively. This indicates that for the unimodal function, as the population size increases, the performance of APO is improved. The average ranking of different population sizes under that function is shown for the multimodal, hybrid, and composition functions. It can be seen that $ps$ 100, and $ps$ 80 perform better, except that $ps$ 80 ranks third in the hybrid function. The ranking is the comparison result of three population sizes under four types of functions. The rankings of large, medium, and small population sizes are first, second, and third. The results show that for our proposed APO algorithm, large populations outperform medium and small populations.

### 3.3. Comparison of APO with some state-of-the-art algorithms metaheuristic algorithms

The proposed APO algorithm was compared with 32 state-of-the-art algorithms under the CEC2022 benchmark. $ps$ was set to 100, max iteration was set to 10,000 ($MaxFEs = 1000,000$), and each benchmark

X. Wang et al.	Knowledge-Based Systems 295 (2024) 111737Table 4
Comparison of the proposed APO algorithm for different population sizes.

|  | Small population | | Medium population | | Large population | |
| --- | --- | --- | --- | --- | --- | --- |
| ps | 10 | 20 | 40 | 50 | 80 | 100 |
| Unimodal function (F1) | 6 | 5 | 3 | 4 | 2 | 1 |
| Multimodal functions (F2–F5) | 6 | 5 | 4 | 3 | 2 | 1 |
| Hybrid functions (F6–F8) | 6 | 4 | 5 | 2 | 3 | 1 |
| Composition functions (F9–F12) | 5 | 6 | 4 | 3 | 2 | 1 |
| **Mean rank (U/M/H/C)** | 5.5/5.5/5/5.5 | | 3.5/3.5/3.5/3.5 | | 1.5/1.5/2/1.5 | |
| **Ranking (U/M/H/C)** | 3/3/3/3 | | 2/2/2/2 | | 1/1/1/1 | |

function was shifted to the same global minimum "0". Table 5 lists the detailed results. "Mean" and "Std" denote the mean and standard deviation of the best-so-far results given by each algorithm, respectively. For enhanced readability, the best "Mean" result is highlighted in light red, while the best "Std" result is highlighted in green. Comparing the mean results, it is found that the proposed APO algorithm is particularly close to the SDO algorithm on F1 and has the best results among the algorithms on F3–F4 and F6–F9. For the remaining test functions, although the proposed APO algorithm did not obtain the best results, it was still superior to most algorithms. The comparison results of 16 advanced algorithms are placed in the supplementary material, refer to Table A.3–A.6. They include algorithms: aquila optimizer (AO) [55], artificial rabbits optimization (ARO) [56], coati optimization algorithm (COA) [57], dandelion optimizer (DO) [58], electric eel foraging optimization (EEFO) [59], five phases algorithm (FPA) [60], gannet optimization algorithm (GOA) [61], hippopotamus optimization (HO) [62], mountain gazelle optimizer (MGO) [63], marine predators algorithm (MPA) [64], quasi-affine transformation evolutionary (QUATRE) [65], reptile search algorithm (RSA) [66], sea-horse optimizer (SHO) [67], slime mould algorithm (SMA) [68], tunicate swarm algorithm (TSA) [69], and zebra optimization algorithm (ZOA) [70].

*3.3.1. Non-parametric statistical analysis*

To better investigate the performance of APO, we applied non-parametric test methods to assess the significant differences among the algorithms. Any two algorithms were compared using the Wilcoxon signed-rank test, whereas multiple algorithms were compared using the Friedman test [71].

Wilcoxon signed-rank test can determine whether there is a statistical difference in the results between the proposed APO algorithm and other algorithms. The test was conducted at the 5% significance level. The statistical results are shown at the end of Table 5. When compared to the other algorithms, "Win", "Draw", and "Lost" indicate that the APO algorithm achieves "better performance", "similar performance", and "worse performance", respectively. In comparison to the GA, Jaya, GWO, WOA, PPE, SCA, AOA, SPO, and SPBO, our proposed APO algorithm can win 12 times, draw 0 times, and lose 0 times. The results indicate that among the compared algorithms, the DE and BSA algorithms exhibit strong performance. However, the proposed APO algorithm still outperforms them, winning 7 times, drawing 3 times, and losing only 2 times. The parameters *p*-value, $R^+$, $R^-$ involved in the Wilcoxon signed-rank test method, and the comparison results of each function can be viewed in Table A.1 in the supplementary material.

Friedman test is a comparison test that can find significant differences among multiple algorithms. The overall ranking of the 17 algorithms on the 12 benchmark functions is shown at the end of Table 5. According to the ranking results, the proposed APO algorithm performed best, ranking first among the 17 algorithms, followed by BSA and DE algorithms. SCA, AOA, and SPO algorithms show worse results. Some of the algorithms showed the same performance, for example, MVO and SPBO both ranked fifth, PSO and TLBO both ranked ninth, and Jaya and WOA both ranked thirteenth. For the Friedman test method, the average ranking of the algorithms on the 12 functions, and the ranking of each function, refer to Table A.2 in the supplementary material. From the above results, the APO evidently outperformed the compared state-of-the-art algorithms on the CEC2022 benchmark. Consequently, the proposed APO algorithm is considered a novel and promising optimization method.

*3.3.2. Convergence analysis*

The convergence of metaheuristic algorithms is closely related to exploration and exploitation. Exploration is an extensive search of the solution space to find potentially better solutions rather than local optima. However, exploitation is a search toward more promising areas, accelerating the discovery of optimal or near-optimal solutions. A balance between exploration and exploitation is crucial to the algorithm's performance. If the algorithm excessively emphasizes exploration, it may waste too much time exploring unknown areas and fail to effectively exploit known information to accelerate convergence toward the optimal solution. This could result in slower convergence or failure to converge to satisfactory solutions. Conversely, If the algorithm excessively exploits known solutions, it may prematurely converge to local optima, preventing exploration of other potential solutions in the search space. This could eventually converge to suboptimal solutions. Therefore, we used a dimension-wise diversity measurement [72,73] to evaluate the performance of the proposed APO algorithm. The exploration and exploitation rates are calculated as follows:

$$err = \frac{div}{div_{max}} \quad (18)$$

$$eir = \frac{div_{max} - div}{div_{max}} \quad (19)$$

$$div = \frac{1}{dim \cdot ps} \cdot \sum_{j=1}^{dim} \sum_{i=1}^{ps} \left| median(x^j) - x_i^j \right| \quad (20)$$

where *err* and *eir* are the exploration and exploitation rates, respectively, for an iteration. $div$ and $div_{max}$ represent the diversity of the entire population in one generation and maximum diversity in all generations, respectively. $median(x^j)$ and $x_i^j$ denote the median value of the *j*th dimension across all individuals and *j*th dimension value of the *i*th individual, respectively.

For exploration and exploitation rates, 100 protozoa were simulated over 500 iterations on the functions (F1–F12). Fig. 7 plots the results obtained by the APO algorithm. In the initial iterations, the proposed APO algorithm emphasizes intensive exploration with limited exploitation. Conversely, as the iterations progress, it transitions to reduced exploration and heightened exploitation. The performance of the algorithm is affected by the test functions, and it can be seen that as the number of iterations increases, the exploration rate gradually decreases, while the exploitation rate gradually increases, and the later exploitation rate is significantly higher than the exploration rate except for the F7 and F8 test functions. The APO algorithm demonstrates good abilities to balance exploration and exploitation during the iterations.

To further explore the convergence of the APO algorithm, we conducted a related experiment: simulating five protozoa in 20 dimensions over 150 iterations. Fig. 8 illustrates the experimental results in two-dimensional space, providing additional insights for the evaluation. The first column shows the fitness landscapes of the unimodal (F1), multimodal (F2–F5), and composition (F9–12) functions. The second column plots the search history, and the red pentagram represents the

**Table 5**
Experimental results of artificial protozoa optimizer and 16 comparative algorithms on CEC2022.

| Fun. | Index | APO | GA | DE | BSA | Jaya | PSO | GWO | WOA | PPE | GSA | MVO | SCA | AOA | TLBO | SDO | SPO | SPBO |
|---|---|---|---|---|---|---|---|---|---|---|---|---|---|---|---|---|---|---|
| F1 | Mean | 1.5158E−14 | 1.7863E+03 | 6.3541E+01 | 2.4828E−04 | 7.4929E+03 | 3.2043E−06 | 4.9282E+03 | 1.9456E+01 | 1.5026E−04 | 9.7105E+03 | 1.7147E−03 | 4.7965E+03 | 3.0441E+04 | 1.5334E+03 | 0.0000E+00 | 8.5843E+04 | 6.7431E+02 |
|  | Std | 2.5567E−14 | 3.1889E+02 | 2.7482E+01 | 5.2037E−04 | 1.4638E+03 | 2.6646E−06 | 3.0532E+03 | 1.9551E+01 | 7.8548E−05 | 1.8891E+03 | 8.3213E−04 | 1.4039E+03 | 8.9421E+03 | 1.1835E+03 | 0.0000E+00 | 3.7505E+04 | 2.2710E+02 |
| F2 | Mean | 4.8910E+01 | 7.3987E+01 | 4.4900E+01 | 4.7533E+01 | 8.9697E+01 | 2.8567E+01 | 7.2824E+01 | 6.1174E+01 | 4.9858E+01 | 4.9279E+01 | 4.3821E+01 | 1.8378E+02 | 2.0569E+03 | 4.6685E+01 | 3.7213E+01 | 1.1913E+03 | 5.3797E+01 |
|  | Std | 9.5627E−01 | 3.9499E+00 | 8.7486E+00 | 7.7468E+00 | 1.8721E+01 | 2.1522E+01 | 2.7904E+01 | 2.3629E+01 | 1.9072E+01 | 1.0647E+00 | 1.1114E+01 | 2.5970E+01 | 7.1438E+02 | 3.5117E+00 | 2.0918E+01 | 5.3192E+02 | 1.2052E+01 |
| F3 | Mean | 7.5791E−14 | 5.3975E+00 | 1.1369E−13 | 1.0611E−13 | 1.3499E+01 | 3.1822E+01 | 9.7599E−01 | 5.2128E+01 | 2.0264E+00 | 3.3315E+01 | 9.7312E−01 | 2.9503E+01 | 5.2497E+01 | 3.0344E+01 | 6.6182E−01 | 4.2912E+01 | 8.0179E−02 |
|  | Std | 5.4509E−14 | 6.0488E−01 | 0.0000E+00 | 2.8843E−14 | 1.6195E+00 | 7.4213E+00 | 8.2626E−01 | 1.4155E+01 | 3.9980E+00 | 7.9191E+00 | 1.9870E+00 | 3.6768E+00 | 4.4944E+00 | 7.1651E+00 | 1.0572E+00 | 1.7939E+01 | 4.7659E−02 |
| F4 | Mean | 5.8080E+00 | 8.9475E+01 | 3.3651E+01 | 2.5623E+01 | 1.1365E+02 | 5.9100E+01 | 4.0720E+01 | 1.0739E+02 | 5.6613E+01 | 7.3925E+01 | 4.2503E+01 | 1.1753E+02 | 1.0365E+02 | 8.2900E+01 | 3.2966E+01 | 1.6059E+02 | 9.6710E+01 |
|  | Std | 1.9764E+00 | 5.5059E+00 | 4.6664E+00 | 5.7189E+00 | 9.0888E+00 | 1.2083E+01 | 2.0575E+01 | 2.5710E+01 | 1.2902E+01 | 8.2633E+00 | 1.7080E+01 | 9.5302E+00 | 1.1227E+01 | 1.9485E+01 | 1.1210E+01 | 3.5923E+01 | 1.0123E+01 |
| F5 | Mean | 2.9843E−03 | 5.4604E+01 | 1.6489E+00 | 1.3642E−13 | 3.8385E+02 | 3.9023E+02 | 6.4314E+01 | 2.2903E+03 | 5.5267E+02 | 0.0000E+00 | 1.0614E−01 | 6.3936E+02 | 1.5253E+03 | 1.4380E+03 | 1.8378E+01 | 2.9671E+03 | 4.9628E+00 |
|  | Std | 1.6346E−02 | 1.4024E+01 | 1.1475E+00 | 4.6252E−14 | 1.1786E+02 | 3.0567E+02 | 5.6484E+01 | 1.1784E+03 | 1.5589E+02 | 0.0000E+00 | 4.0878E−01 | 1.9671E+02 | 7.1773E+01 | 6.0650E+02 | 2.4265E+01 | 2.0524E+03 | 4.9693E+00 |
| F6 | Mean | 3.5179E+01 | 1.6627E+03 | 7.6174E+03 | 4.6660E+01 | 2.8479E+07 | 4.1492E+02 | 5.4369E+04 | 5.6291E+03 | 2.2926E+03 | 1.1456E+03 | 1.3747E+04 | 4.7086E+07 | 1.1610E+05 | 6.2156E+04 | 1.1360E+02 | 5.4073E+08 | 4.7675E+03 |
|  | Std | 2.2617E+01 | 9.9000E+02 | 3.9527E+03 | 2.0440E+01 | 1.4237E+07 | 7.8275E+02 | 1.6863E+05 | 5.8590E+03 | 1.7217E+03 | 6.6349E+02 | 8.9344E+03 | 2.9946E+07 | 6.2465E+05 | 1.6875E+05 | 1.0753E+02 | 8.2470E+08 | 4.2408E+03 |
| F7 | Mean | 1.2911E+01 | 3.8782E+01 | 2.0813E+01 | 1.3914E+01 | 9.4879E+01 | 8.9918E+01 | 4.8326E+01 | 1.4679E+02 | 5.2165E+01 | 3.1833E+02 | 4.3184E+01 | 9.7681E+01 | 1.7715E+02 | 8.4191E+01 | 4.1962E+01 | 1.7378E+02 | 3.2480E+01 |
|  | Std | 7.5497E+00 | 8.9573E+00 | 3.6168E+00 | 6.4545E+00 | 9.4474E+00 | 4.2943E+01 | 2.5962E+01 | 4.7372E+01 | 1.4628E+01 | 1.4628E+01 | 2.8475E+01 | 1.3681E+01 | 1.5755E+01 | 1.3366E+01 | 6.0390E+01 | 6.3137E+00 |
| F8 | Mean | 1.9754E+01 | 2.6203E+01 | 2.2154E+01 | 2.1243E+01 | 3.5182E+01 | 5.9918E+01 | 3.8631E+01 | 4.1587E+01 | 3.4610E+01 | 2.4136E+01 | 6.9278E+01 | 4.1606E+01 | 1.2427E+02 | 3.4974E+01 | 2.1816E+01 | 9.7509E+01 | 3.3369E+01 |
|  | Std | 2.3634E+00 | 8.5478E−01 | 4.0658E−01 | 1.3709E+00 | 3.0255E+00 | 5.5722E+01 | 5.5733E+01 | 1.1228E+01 | 3.5987E+01 | 5.4998E+01 | 6.8399E+01 | 3.3712E+00 | 7.6430E+01 | 3.0702E+00 | 1.0542E+00 | 4.2687E+03 | 1.2634E+00 |
| F9 | Mean | 1.8078E+02 | 1.8933E+02 | 1.8078E+02 | 1.8078E+02 | 1.8977E+02 | 1.8934E+02 | 1.9164E+02 | 1.8119E+02 | 1.8080E+02 | 1.8078E+02 | 1.8079E+02 | 2.2497E+02 | 6.3915E+02 | 1.8089E+02 | 1.8078E+02 | 5.1082E+02 | 1.8079E+02 |
|  | Std | 8.6723E−14 | 1.2939E+00 | 3.7513E−05 | 1.6471E−13 | 2.3377E+00 | 1.9528E+01 | 1.5930E+01 | 4.3189E−01 | 1.5431E+00 | 1.2820E−13 | 6.6539E−03 | 1.4006E+00 | 1.3441E+02 | 6.8565E−02 | 8.6723E−14 | 1.6157E+03 | 4.0761E−03 |
| F10 | Mean | 1.0033E+02 | 1.2860E+02 | 9.3867E+01 | 1.0049E+02 | 1.3753E+03 | 1.0626E+03 | 6.8868E+02 | 1.5229E+03 | 2.2946E+02 | 2.4475E+03 | 1.2204E+03 | 1.1273E+02 | 1.5317E+03 | 1.3511E+02 | 1.0059E+02 | 2.6770E+03 | 1.4791E+02 |
|  | Std | 3.6067E−02 | 7.1964E+01 | 2.4752E+01 | 4.7469E−02 | 1.3769E+03 | 9.4515E+02 | 5.4130E+02 | 1.0673E+03 | 1.6348E+02 | 5.6397E+02 | 8.6706E+02 | 3.4417E+00 | 5.4484E+02 | 1.5561E+02 | 1.3271E−01 | 1.2328E+03 | 7.4444E+01 |
| F11 | Mean | 3.0333E+02 | 6.8170E+02 | 2.8369E+02 | 1.6002E+02 | 1.0729E+03 | 3.5370E+02 | 6.5005E+02 | 5.2810E+02 | 3.2001E+02 | 3.5000E+02 | 3.3799E+02 | 1.4128E+02 | 5.8495E+03 | 3.0510E+02 | 3.2667E+02 | 4.5123E+03 | 3.5210E+02 |
|  | Std | 1.8257E+01 | 3.4694E+01 | 1.3691E+02 | 1.4344E+02 | 2.1041E+02 | 1.6778E+02 | 1.9816E+02 | 1.0462E+03 | 7.6107E+01 | 5.0855E+01 | 7.1116E+01 | 3.4108E+02 | 8.6153E+02 | 1.5024E+02 | 4.4978E+01 | 1.4030E+03 | 8.3403E+01 |
| F12 | Mean | 2.3711E+02 | 2.7515E+02 | 2.3503E+02 | 2.3392E+02 | 2.4847E+02 | 3.8741E+02 | 2.5356E+02 | 3.2229E+02 | 4.4787E+02 | 5.0753E+02 | 2.4295E+02 | 2.9651E+02 | 7.4006E+02 | 2.6070E+02 | 2.6416E+02 | 5.1297E+02 | 2.4515E+02 |
|  | Std | 3.6438E+00 | 6.4562E+00 | 1.6015E+00 | 1.8457E+00 | 2.6293E+00 | 1.3224E+02 | 1.3748E+01 | 6.9277E+01 | 8.2984E+01 | 1.2039E+02 | 9.7241E+00 | 1.0365E+01 | 1.3497E+02 | 1.7575E+01 | 1.3774E+01 | 1.2775E+03 | 8.4873E+00 |
| Win/Draw/Lost |  | compared | 12/0/0 | 7/3/2 | 7/3/2 | 12/0/0 | 11/0/1 | 12/0/0 | 12/0/0 | 12/0/0 | 10/0/2 | 11/0/1 | 12/0/0 | 12/0/0 | 11/0/1 | 9/1/2 | 12/0/0 | 12/0/0 |
| Ranking |  | 1 | 8 | 3 | 2 | 13 | 9 | 11 | 13 | 7 | 12 | 5 | 15 | 16 | 9 | 4 | 17 | 5 |

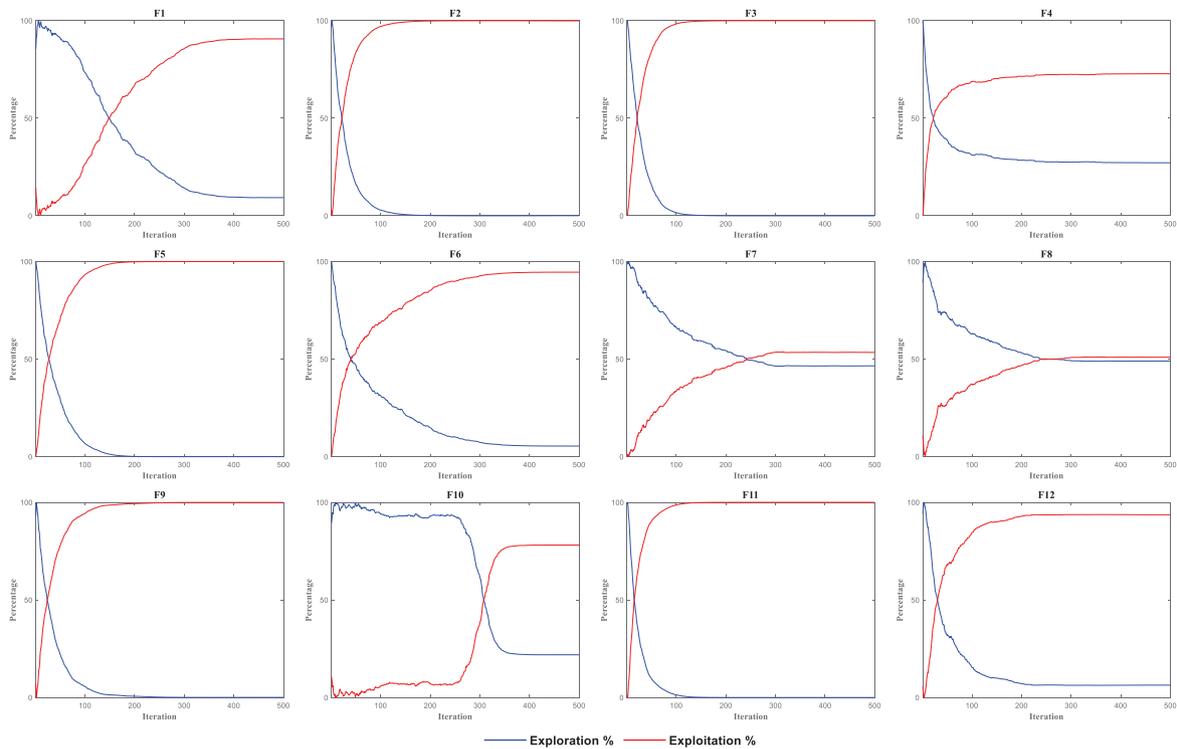

**Fig. 7.** Exploration and exploitation rate for F1–F12.

obtained global optimum. From the sparsity of the sample points, we can observe that the distribution of points far from the global optimum is sparse, whereas the distribution of points close to it is dense. This shows that the APO can achieve extensive exploration of space and intensive exploitation of promising areas. The third and fourth columns show the trajectories in the first and second dimensions of the five protozoa over the iterations, respectively. In the previous iterations, the values of the protozoa changed drastically and abruptly, but this change gradually faded in later iterations. The fifth column shows the average fitness values of the five protozoa. The average fitness value gradually decreases throughout the iteration process, signaling an improvement in the overall quality of the candidate solutions. The sixth column presents the convergence curves. Throughout each iteration, the fitness value of the best protozoan gradually decreases, indicating that the APO algorithm gradually converges. In summary, these indicators demonstrate that our APO algorithm effectively maintains a balance between the exploration and exploitation phases, leading to an improved solution quality within the population.

The convergence behavior of the proposed APO algorithm is analyzed above. Next, the convergence of the APO algorithm is compared with the state-of-the-art algorithms in the unimodal, multimodal, hybrid, and composition functions. Unimodal functions have a sole locally optimal solution, serving as a benchmark for evaluating the algorithm's exploitation capability. Conversely, multimodal functions contain multiple locally optimal solutions and are evaluated for the exploration of the algorithm. Hybrid functions are a combination of unimodal and multimodal functions and present a more complex search space structure. Composition functions are complex functions consisting of multiple simple functions, entailing a highly nonlinear and irregular search space. Hybrid and composition functions can assess the balance between exploration and exploitation. The mean fitness convergence curves of the algorithms are shown in Fig. 9. For the unimodal function F1, SDO performs best and can find the global optimal solution 0. The proposed APO algorithm is also very competitive, achieving the result with a mean fitness value of 1.5158E−14, and it converges faster. These two algorithms are significantly better than other comparison algorithms. For multimodal functions, on the F3 and F4 functions, the proposed APO performs best among the algorithms and converges faster. The results on F2 and F5 are also better than most algorithms. For hybrid functions (F6–F8), the proposed APO performs better than all comparison algorithms. It can be seen that the comparison algorithms are premature or have not converged. Moving to the composition functions, the proposed APO still performs best on the F9 function, and DE, BSA, GSA, and SDO also show similar results. For the F10 function, the proposed APO performs only worse than DE. For F11 and F12, the BSA algorithm performs best, followed by DE and our proposed APO algorithm. Convergence analysis confirmed that the APO algorithm can perform better solutions with faster convergence than its compared algorithms in most benchmark functions.

## 4. Application

This section discusses certain practical applications of the proposed APO algorithm in engineering design and multilevel image segmentation problems.

### 4.1. Engineering design

The proposed APO algorithm addressed the design challenges in tension/compression springs, pressure vessels, welded beams, speed reducers, and three-bar trusses [74–76]. Fig. 10 illustrates five widely recognized engineering designs, with specific details available in the supplementary material (I-V). These five engineering designs are continuous space optimization problems with constraints. Various strategies are discussed for handling constraints in [77,78], including the incorporation of penalty functions, decoder functions, repair algorithms, special operators, feasibility-preserving representations, and the separation of objectives and constraints. In this study, the penalty function was chosen because of its simplicity and low computational cost. It directly penalizes infeasible candidate solutions and then converts constrained optimization into unconstrained optimization. For minimization problems, the penalty function assigns large function values to infeasible solutions, which causes them to be discarded by metaheuristic algorithms during optimization. In the experiments, *ps* was set to 100, the maximum number of iterations was set to 500, and the number of runs for the algorithms was 31; for the other parameters, refer to Section 3.1. We selected the first half of the algorithms in Table 3 for comparison (evolutionary- and swarm-based algorithms). The optimal results obtained for each design problem are highlighted in bold font in the tables.

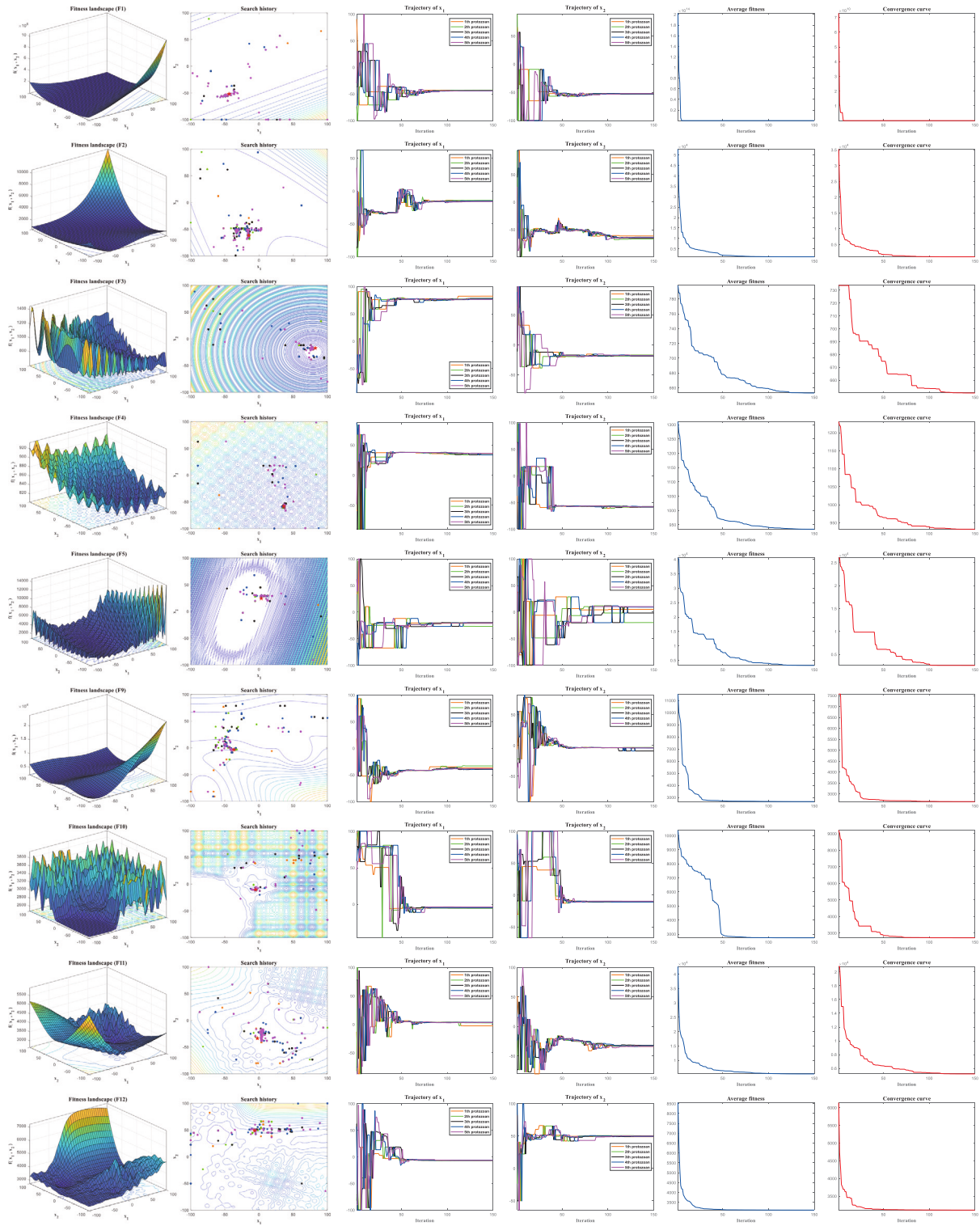

**Fig. 8.** Fitness landscape, search history, trajectory, average fitness, and convergence curve of unimodal (F1), multimodal (F2–F5), and composition (F9–12) functions.

*4.1.1. Tension/compression spring*

The goal of the tension/compression spring design is to minimize the manufacturing costs by optimizing three structural variables: the wire diameter ($d$), mean coil diameter ($D$), and number of active coils ($N$). The design process is governed by constraints related to the shear stress, surge frequency, and minimum deflection. The best experimental results of the APO and comparative algorithms are listed in Table 6.

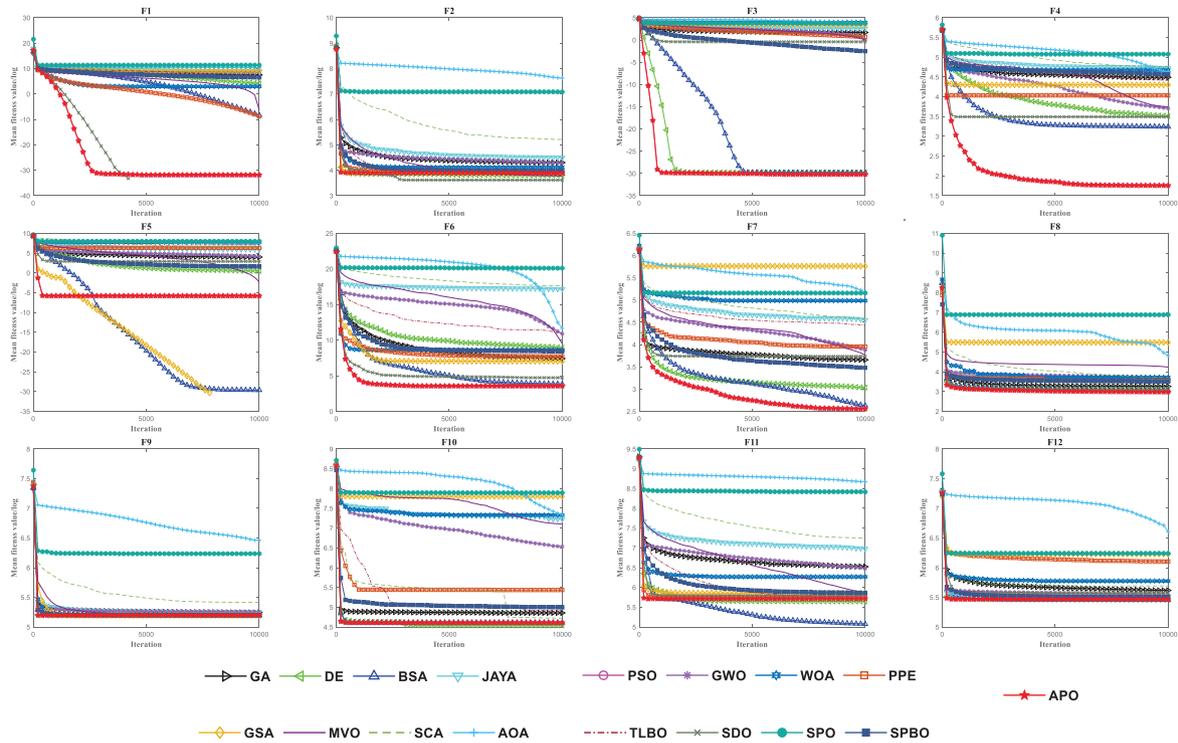

**Fig. 9.** Convergence curves of the mean fitness value of F1–F12.

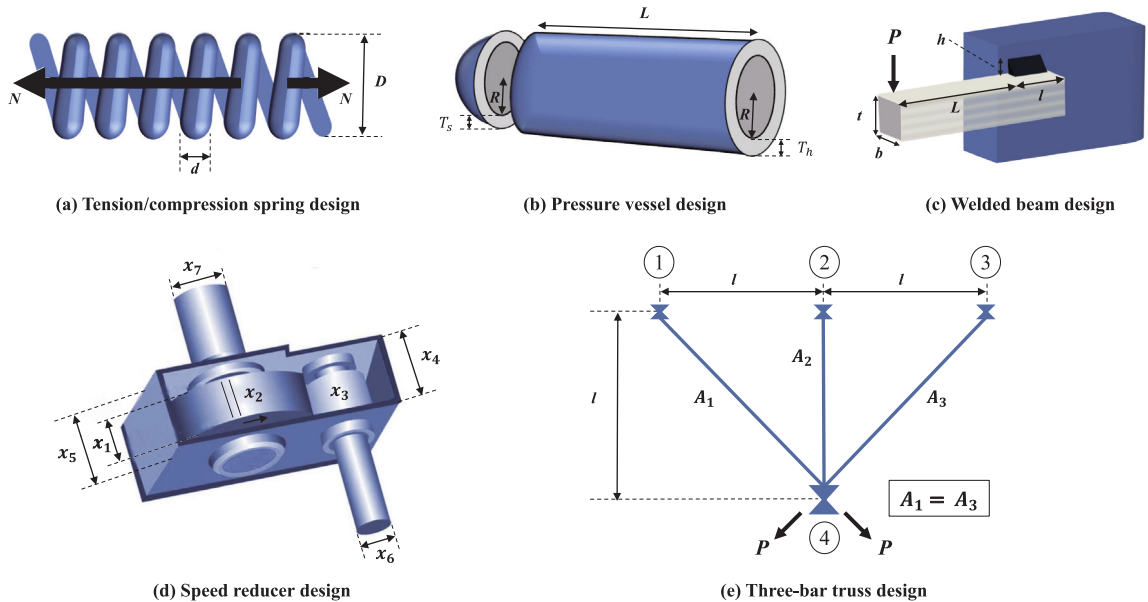

(a) Tension/compression spring design

(b) Pressure vessel design

(c) Welded beam design

(d) Speed reducer design

(e) Three-bar truss design

**Fig. 10.** Five engineering design problems.

Notably, the APO achieves the lowest cost (0.01266529) under specific variables (0.0516521, 0.355829, and 11.3413). When compared with the eight algorithms (GA, DE, BSA, Jaya, PSO, GWO, WOA, and PPE), the proposed algorithm demonstrates greater competitiveness in addressing the tension/compression spring design problem.

*4.1.2. Pressure vessel*
The pressure vessel design aims to minimize the overall costs related to materials, forming, and welding. This design challenge involves determining the shell thickness ($T_s$), head thickness ($T_h$), inner radius ($R$), and cylindrical section length ($L$) while satisfying four specified constraints. The best experimental results are listed in Table 7. When the variables were set at 0.77916, 0.38516, 40.3707, and 199.3144, the proposed APO algorithm achieved an optimal cost of 5887.614. The APO algorithm consistently outperforms its counterparts in the pressure vessel design problem, excluding Jaya and PSO, offering more competitive results.

*4.1.3. Welded beam*
The objective of designing a welded beam is to minimize the expenses related to manufacturing and optimize four crucial structural variables: weld thickness ($h$), length of clamped steel bars ($l$), height of steel bars ($t$), and thickness of bars ($b$). This design process is constrained by the shear stress ($\tau$), bending stress in the beam ($\sigma$), buckling load on the bar ($P_c$), end deflection of the beam ($\delta$), and side constraints. The challenge is to achieve an optimal solution for these variables within the specified constraints, ensuring both cost effectiveness and structural reliability of the welded beam. Table 8 lists the best experimental outcomes generated by the algorithms. In this design problem, the APO algorithm achieves an optimal cost (1.724854) with the variables (0.20573, 3.4705, 9.0366, and 0.20573), mirroring the results obtained by Jaya and PSO. These results highlighted the effectiveness of the proposed APO in the welded-beam design problem.

**Table 6**
Tension/compression spring design: the experimental results of APO and eight comparative algorithms.

| Algorithm | Optimal variables | | | Optimum |
|---|---|---|---|---|
| | $d$ | $D$ | $N$ | |
| APO | 0.0516521 | 0.355829 | 11.3413 | **0.01266529** |
| GA | 0.0603490 | 0.599940 | 4.55340 | 0.01459647 |
| DE | 0.0521916 | 0.368916 | 10.6088 | 0.01267075 |
| BSA | 0.0520386 | 0.365182 | 10.8110 | 0.01266898 |
| Jaya | 0.0514321 | 0.350519 | 11.6747 | 0.01267934 |
| PSO | 0.0517542 | 0.358286 | 11.1976 | 0.01266531 |
| GWO | 0.0518078 | 0.359533 | 11.1281 | 0.01266866 |
| WOA | 0.0514827 | 0.351773 | 11.5848 | 0.01266602 |
| PPE | 0.0524206 | 0.374571 | 10.3145 | 0.01267525 |

**Table 7**
Pressure vessel design: the experimental results of APO and eight comparative algorithms.

| Algorithm | Optimal variables | | | | Optimum |
|---|---|---|---|---|---|
| | $T_s$ | $T_h$ | $R$ | $L$ | |
| APO | 0.77916 | 0.38516 | 40.3707 | 199.3144 | 5887.614 |
| GA | 1.14157 | 0.57814 | 58.7532 | 42.0918 | 6998.416 |
| DE | 0.78597 | 0.38862 | 40.7128 | 194.6202 | 5901.043 |
| BSA | 0.78054 | 0.38976 | 40.4390 | 198.5150 | 5904.985 |
| Jaya | 0.77817 | 0.38465 | 40.3196 | 200 | **5885.336** |
| PSO | 0.77817 | 0.38482 | 40.3196 | 200 | 5885.813 |
| GWO | 0.77946 | 0.38549 | 40.3836 | 199.1158 | 5888.614 |
| WOA | 0.79192 | 0.38838 | 40.6099 | 195.9977 | 5956.492 |
| PPE | 0.83161 | 0.41430 | 43.0675 | 165.1467 | 6000.281 |

*4.1.4. Speed reducer*

The primary goal in the design of a speed reducer is weight minimization. This design process considers various parameters, including surface width ($x_1$), tooth module ($x_2$), number of teeth in the pinion ($x_3$), length of first shaft between bearings ($x_4$), length of second shaft between bearings ($x_5$), diameter of the first shaft ($x_6$), and diameter of the second shaft ($x_7$). Moreover, this design is bound by constraints such as managing the bending stresses in the gear teeth, surface stresses, lateral deflection of the shaft, and stresses within the shaft. With respect to this design challenge, Table 9 shows that the APO contributes the most favorable weight (2994.471) using different variables (3.5, 0.7, 17, 7.3, 7.71552, 3.35021, 5.28665). DE and Jaya also demonstrate comparable performance. When addressing this design problem, the APO outperformed the comparison algorithms.

*4.1.5. Three-bar truss*

The purpose of the three-bar truss design is weight minimization. This design involves two variables: the cross-sectional areas of the first ($A_1$) and second ($A_2$) bars. The design was subjected to constraints encompassing the stress, deflection, and buckling of each bar. Table 10 lists the best experimental results from each algorithm. Notably, the APO achieves the optimal weight (263.8958) with the corresponding variables (0.78868 and 0.40825). Furthermore, DE and PSO also demonstrate commendable performances. The experimental results indicate the effectiveness of the proposed APO algorithm for this design problem.

Table 11 provides an overview of the best results for the nine algorithms. Notably, the APO stands out as the top-ranking algorithm for the tension/compression spring design. PSO ranks second, WOA ranks third, GWO ranks fourth, BSA ranks fifth, DE ranks sixth, PPE ranks seventh, Jaya ranks eighth, and GA ranks ninth. For the rankings of the other four design problems, refer to that of the tension/compression spring design problem. According to the Friedman test method, the overall ranking of the five designs is as follows: APO ranks first, PSO ranks second, Jaya ranks third, DE ranks fourth, GWO ranks fifth, BSA ranks sixth, PPE ranks seventh, WOA ranks eighth, and GA ranks ninth.

*4.1.6. Analysis of stability and computational complexity on the five engineering design problems*

The performance of metaheuristic algorithms is affected by problem complexity, and they inherently operate as stochastic methods. Analyzing algorithm stability and computational efficiency is essential. This part of the research investigates the stability and computational complexity of the proposed algorithm that performs on five engineering design problems. The task of the algorithms is to find feasible solutions for the engineering design problems in the minimum time. Three evaluation indicators are introduced: success rate ($SR$), average number of fitness evaluations ($AFEs$), and average calculation durations ($ACDs$) [79], defined as follows:

$$SR = \frac{r}{R} \cdot 100 \tag{21}$$

$$AFEs = \frac{1}{r} \cdot \sum_{i=1}^{r} FEs(i) \tag{22}$$

$$ACDs = \frac{1}{r} \cdot \sum_{i=1}^{r} Calculation\_Duration(i) \tag{23}$$

where $r$ is the number of successful searches for a feasible solution defined by the problem. $R$ is the number of algorithm runs. $FEs$ and $Calculation\_Duration$ are the number of fitness evaluations, and the running time used by the algorithm to find the feasible solution.

According to the mean result of each algorithm, this paper selects the median ranking algorithm as the acceptable standard, and its mean result is used as a feasible solution to each engineering design problem. For example, in the tension/compression spring problem, Jaya ranks fifth among the nine algorithms, and its mean value is defined as the feasible solution to this problem. The feasible solutions to the other four engineering design problems are selected from the algorithms APO, DE, PPE, and PPE respectively. Table 12 gives feasible solutions to five engineering design problems. Success rate, average number of fitness evaluations, and average calculation durations are presented in Table 13. The comparison of algorithms first considers the success rate, and then compares the average number of fitness evaluations, and average calculation durations. The best results obtained by the algorithms have been bolded. For the tension/compression spring design, the proposed APO algorithm searches for the feasible solution with *SR* of 100%, *AFEs* of 9900, and *ACDs* of 0.13 s, which outperforms the other eight algorithms. From the results of five engineering design problems, the proposed APO algorithm achieves 100% *SR* on the four problems, and its *AFEs* and *ACDs* are also competitive. This result demonstrates the advantages of the APO algorithm's stability and computational efficiency. For the pressure vessel problem, the performance of APO is worse than Jaya, DE, BSA, and GWO, but better than PSO, PPE, WOA, and GA. It is worth mentioning that the Jaya algorithm closely follows the proposed APO algorithm, which obtains 100% *SR* with high computational efficiency on the three problems. GA has the worst performance among the algorithms. It failed to find feasible solutions for the five problems. In summary, the above experiments confirm that the proposed APO algorithm outperforms the eight state-of-the-art algorithms in solving the five engineering design problems.

**Table 8**
Welded-beam design: the experimental results of APO and eight comparative algorithms.

| Algorithm | Optimal variables | | | | Optimum |
|---|---|---|---|---|---|
| | h | l | t | b | |
| APO | 0.20573 | 3.4705 | 9.0366 | 0.20573 | **1.724854** |
| GA | 0.23614 | 3.1413 | 8.3938 | 0.24180 | 1.867287 |
| DE | 0.20553 | 3.4752 | 9.0413 | 0.20666 | 1.733063 |
| BSA | 0.20518 | 3.4599 | 9.0963 | 0.20552 | 1.731286 |
| Jaya | 0.20573 | 3.4705 | 9.0366 | 0.20573 | **1.724854** |
| PSO | 0.20573 | 3.4705 | 9.0366 | 0.20573 | **1.724854** |
| GWO | 0.20562 | 3.4748 | 9.0374 | 0.20573 | 1.725369 |
| WOA | 0.20166 | 3.5605 | 9.0362 | 0.20687 | 1.739211 |
| PPE | 0.20560 | 3.4744 | 9.0366 | 0.20574 | 1.725258 |

**Table 9**
Speed reducer design: the experimental results of APO and eight comparative algorithms.

| Algorithm | Optimal variables | | | | | | | Optimum |
|---|---|---|---|---|---|---|---|---|
| | $x_1$ | $x_2$ | $x_3$ | $x_4$ | $x_5$ | $x_6$ | $x_7$ | |
| APO | 3.5 | 0.7 | 17 | 7.3 | 7.71532 | 3.35021 | 5.28665 | **2994.471** |
| GA | 3.51538 | 0.70061 | 17 | 7.35568 | 7.74519 | 3.36023 | 5.28932 | 3012.430 |
| DE | 3.5 | 0.7 | 17 | 7.3 | 7.71532 | 3.35021 | 5.28665 | **2994.471** |
| BSA | 3.5 | 0.7 | 17 | 7.3 | 7.71534 | 3.35022 | 5.28666 | 2994.472 |
| Jaya | 3.5 | 0.7 | 17 | 7.3 | 7.71532 | 3.35021 | 5.28665 | **2994.471** |
| PSO | 3.5 | 0.7 | 17 | 8.26743 | 8.3 | 3.35214 | 5.28686 | 3016.465 |
| GWO | 3.50075 | 0.7 | 17 | 7.37393 | 7.71758 | 3.36224 | 5.28666 | 2998.546 |
| WOA | 3.50263 | 0.7 | 17 | 7.68301 | 7.72017 | 3.35334 | 5.28666 | 2999.791 |
| PPE | 3.50022 | 0.70002 | 17 | 7.30673 | 7.71619 | 3.35034 | 5.28669 | 2994.818 |

**Table 10**
Three-bar truss design: the experimental results of APO and eight comparative algorithms.

| Algorithm | Optimal variables | | Optimum |
|---|---|---|---|
| | $A_1$ | $A_2$ | |
| APO | 0.78868 | 0.40825 | **263.8958** |
| GA | 0.79037 | 0.40347 | 263.8983 |
| DE | 0.78868 | 0.40825 | **263.8958** |
| BSA | 0.78870 | 0.40818 | 263.8960 |
| Jaya | 0.78866 | 0.40828 | 263.8959 |
| PSO | 0.78868 | 0.40825 | **263.8958** |
| GWO | 0.78883 | 0.40781 | 263.8960 |
| WOA | 0.78933 | 0.40639 | 263.8962 |
| PPE | 0.78862 | 0.40840 | 263.8959 |

### 4.2. Multilevel image segmentation

Image segmentation technology, which is an important part of image processing, pattern recognition, and computer vision, provides significant inspiration for subsequent image analyses. Its current applications span diverse fields, including face recognition, video surveillance, object tracking, medical image analysis, image compression, and augmented reality [80–83]. Thresholding is one of the most popular and fundamental approaches in the field of image segmentation. In general, this method selects thresholds according to gray-level histograms and then divides the pixels into different classes. Bi-level image segmentation is a relatively straightforward process that divides an image into objects and backgrounds based on a single threshold. However, as we move to multilevel image segmentation, the complexity of the problem intensifies, and the accuracy tends to decrease because of the increasing number of categories. Consequently, multilevel image segmentation is more challenging. Currently, numerous methods utilizing Renyi [84], Masi [85], Tsallis [86], Shannon [87], and cross- [88–90] entropies have been investigated. In this study, the minimum cross-entropy threshold (MCET) was used as the objective function to calculate the optimal segmentation thresholds for color images.

#### 4.2.1. Minimum cross-entropy threshold

Kullback first presented cross-entropy in [91]. Let $P = \{p_1, p_2, \ldots, p_n\}$ and $Q = \{q_1, q_2, \ldots, q_n\}$ be expressed as two probability distributions. The cross-entropy is an information-theoretic distance between two distributions and is defined as follows:

$$C(P, Q) = \sum_{i=1}^{n} p_i \log \frac{p_i}{q_i} \qquad (24)$$

Multilevel image segmentation focuses on identifying the boundaries to divide an image into multiple regions. Consider dividing an image into $(n + 1)$ classes using $n$ thresholds. The $n$ thresholds are denoted as $\{t_1, t_2, \ldots, t_n\}$. Pixels belonging to $\{0, \ldots, t_1\}$ are in $class_1$, pixels belonging to $\{t_1, \ldots, t_2\}$ are in $class_2$, $\cdots$, and pixels belonging to $\{t_n, \ldots, L\}$ are in $class_{n+1}$. $L$ denotes the maximum value among the pixels. Consequently, the MCET method calculates $n$ thresholds as follows:

$$\{t_1^*, t_2^*, \ldots, t_n^*\} = \arg\min\{f(t_1, t_2, \ldots, t_n)\} \quad (25)$$

$$f(t_1, t_2, \ldots, t_n) = \sum_{i=1}^{L} i \cdot z(i) \cdot \log(i) - \sum_{k=1}^{n+1} \sum_{i=t_{k-1}}^{t_k - 1} i \cdot z(i) \cdot \log(u(t_{k-1}, t_k)), \quad (26)$$

Table 11  
Ranking of algorithms on the five engineering design problems.

|  | APO | GA | DE | BSA | Jaya | PSO | GWO | WOA | PPE |
|---|---|---|---|---|---|---|---|---|---|
| Tension/compression spring | 1 | 9 | 6 | 5 | 8 | 2 | 4 | 3 | 7 |
| Pressure vessel | 3 | 9 | 5 | 6 | 1 | 2 | 4 | 7 | 8 |
| Welded beam | 2 | 9 | 7 | 6 | 2 | 2 | 5 | 8 | 4 |
| Speed reducer | 2 | 8 | 2 | 4 | 2 | 9 | 6 | 7 | 5 |
| Three-bar truss | 2 | 9 | 2 | 6.5 | 4.5 | 2 | 6.5 | 8 | 4.5 |
| **Mean rank** | 2 | 8.8 | 4.4 | 5.5 | 3.5 | 3.4 | 5.1 | 6.6 | 5.7 |
| **Ranking** | **1** | **9** | **4** | **6** | **3** | **2** | **5** | **8** | **7** |

Table 12  
Feasible solutions for the five engineering design problems.

|  | Tension/compression spring | Pressure vessel | Welded beam | Speed reducer | Three-bar truss |
|---|---|---|---|---|---|
| Feasible solutions | 0.01273171 | 6117.896 | 1.765002 | 2996.868 | 263.8963 |

Table 13  
Success rate (SR), average number of fitness evaluations (AFEs) and average calculation durations (ACDs) on the five engineering design problems.

|  |  | APO | GA | DE | BSA | Jaya | PSO | GWO | WOA | PPE |
|---|---|---|---|---|---|---|---|---|---|---|
| Tension/compression spring | SR (%) | **100** | 0 | 67.74 | 80.65 | 96.77 | 83.87 | 96.77 | 16.13 | 6.45 |
|  | AFEs | **9900** | – | 23 267 | 33 684 | 19 447 | 10 819 | 46 517 | 1980 | 9500 |
|  | ACDs (s) | 0.13 | – | 0.24 | 0.15 | 0.10 | 0.04 | 0.24 | 0.01 | 0.68 |
| Pressure vessel | SR (%) | 0.7419 | 0 | 96.77 | 96.77 | **100** | 41.94 | 87.10 | 3.23 | 9.68 |
|  | AFEs | 21 435 | – | 15 703 | 31 880 | **12 042** | 33 254 | 26 248 | 4500 | 40 500 |
|  | ACDs (s) | 0.27 | – | 0.16 | 0.13 | **0.06** | 0.12 | 0.13 | 0.03 | 2.91 |
| Welded beam | SR (%) | **100** | 0 | 54.84 | 74.19 | **100** | 77.42 | **100** | 3.23 | 35.48 |
|  | AFEs | 11 255 | – | 33 847 | 40 600 | **10 394** | 12 604 | 34 565 | 15 000 | 18 309 |
|  | ACDs (s) | 0.15 | – | 0.38 | 0.21 | **0.06** | 0.06 | 0.21 | 0.08 | 1.31 |
| Speed reducer | SR (%) | **100** | 0 | **100** | **100** | **100** | 0 | 0 | 0 | 48.39 |
|  | AFEs | 6148 | – | **5035** | 17 126 | 5642 | – | – | – | 37 767 |
|  | ACDs (s) | 0.11 | – | 0.07 | 0.12 | **0.04** | – | – | – | 2.81 |
| Three-bar truss | SR (%) | **100** | 0 | **100** | **100** | 51.61 | **100** | 12.90 | 3.23 | 61.29 |
|  | AFEs | **6423** | – | 17 155 | 29 216 | 35 538 | 14 916 | 49 200 | 28 800 | 43 089 |
|  | ACDs (s) | 0.08 | – | 0.17 | 0.11 | 0.15 | **0.05** | 0.20 | 0.12 | 3.07 |

Subject to $\quad 0 < t_1 < t_2 < \cdots < t_n < L$

$$u(t_{k-1}, t_k) = \frac{\sum_{i=t_{k-1}}^{t_k - 1} i \cdot z(i)}{\sum_{i=t_{k-1}}^{t_k - 1} z(i)} \quad (27)$$

Here, $t_0 = 1$, $t_{n+1} = L + 1$, $t_i$ denotes the $i$th threshold, and $z_i$ denotes the number of $i$th pixels. The pixels of the original image in the interval $[t_{k-1}, t_k]$ can be calculated as $u(t_{k-1}, t_k)$ after solving for the thresholds.

*4.2.2. Algorithm comparison experiments on multilevel image segmentation*

The performance of the APO was evaluated using multilevel image segmentation. A Lena color image of 512 × 512 pixels was tested. Regarding the experimental parameter settings, $ps$ was 100, the maximum number of iterations was 100, and the number of runs for the algorithms was 31; for the remaining parameters, refer to Section 3.1.

Image segmentation is a discrete optimization method in which each candidate solution is rounded at each iteration for solving this problem. The colored Lena image contains three pixel channels: red, green, and blue. In each channel, the histogram information of the pixels was used to solve the thresholds separately using the APO. Finally, the segmentation results of the three channels were concatenated to obtain the segmented Lena. Fig. 11 illustrates the original Lena image along with the five segmentation cases (2, 4, 6, 8, and 10 thresholds). Noticeably, the quality of the segmented images improves with an increase in the number of thresholds. We utilized three quantitative indicators to evaluate the segmented images: peak signal-to-noise ratio (PSNR) [92], structural similarity index (SSIM) [93], and feature similarity index (FSIM) [94–96]. The PSNR quantifies the peak error, SSIM evaluates structural similarity, and FSIM measures feature similarity. A higher value of these three indicators implies that a better image quality has been segmented by the algorithms.

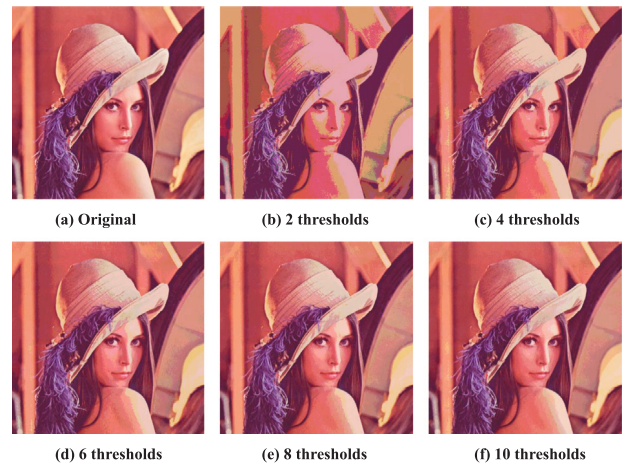

(a) Original  (b) 2 thresholds  (c) 4 thresholds  
(d) 6 thresholds  (e) 8 thresholds  (f) 10 thresholds

**Fig. 11.** Original and segmented Lena images obtained by artificial protozoa optimizer.

Table 14
Comparison of algorithms based on the peak signal-to-noise ratio, structured similarity index, and feature similarity index.

| Threshold | APO | GSA | MVO | SCA | AOA | TLBO | SDO | SPO | SPBO |
|---|---|---|---|---|---|---|---|---|---|
| | | | | | **PSNR** | | | | |
| 2 | **23.5625** | 23.5625 | 23.5625 | 23.5625 | 23.5625 | 23.5625 | 23.5625 | 23.5625 | 23.5625 |
| 4 | **28.4418** | 28.3012 | 28.4418 | 28.4251 | 28.4418 | 28.4325 | 28.4418 | 28.4418 | 28.4418 |
| 6 | 30.9630 | 30.3931 | 30.9804 | 30.7255 | 30.9693 | 30.9313 | 30.9572 | 30.9804 | **30.9842** |
| 8 | **33.0393** | 32.2141 | 33.0292 | 32.4045 | 33.0256 | 32.8765 | 32.9946 | 33.0292 | 32.5281 |
| 10 | **34.6564** | 33.4351 | 34.3325 | 33.7262 | 34.3368 | 34.0703 | 34.1887 | 34.3319 | 34.3698 |
| | | | | | **SSIM** | | | | |
| 2 | **0.8106** | 0.8106 | 0.8106 | 0.8106 | 0.8106 | 0.8106 | 0.8106 | 0.8106 | 0.8106 |
| 4 | 0.8935 | 0.8884 | 0.8935 | **0.8938** | 0.8935 | 0.8936 | 0.8935 | 0.8935 | 0.8935 |
| 6 | 0.9307 | 0.9252 | 0.9308 | 0.9255 | 0.9309 | 0.9304 | 0.9306 | 0.9308 | **0.9314** |
| 8 | **0.9534** | 0.9456 | 0.9533 | 0.9439 | 0.9534 | 0.9513 | 0.9531 | 0.9533 | 0.9473 |
| 10 | **0.9646** | 0.9559 | 0.9641 | 0.9565 | 0.9642 | 0.9621 | 0.9631 | 0.9641 | 0.9632 |
| | | | | | **FSIM** | | | | |
| 2 | **0.8440** | 0.8440 | 0.8440 | 0.8440 | 0.8440 | 0.8440 | 0.8440 | 0.8440 | 0.8440 |
| 4 | 0.9155 | **0.9161** | 0.9155 | 0.9155 | 0.9155 | 0.9153 | 0.9155 | 0.9155 | 0.9155 |
| 6 | 0.9585 | 0.9523 | 0.9584 | **0.9614** | 0.9580 | 0.9570 | 0.9575 | 0.9584 | 0.9588 |
| 8 | **0.9728** | 0.9596 | 0.9727 | 0.9712 | 0.9727 | 0.9724 | 0.9725 | 0.9727 | 0.9702 |
| 10 | **0.9826** | 0.9766 | 0.9790 | 0.9780 | 0.9789 | 0.9787 | 0.9783 | 0.9790 | 0.9803 |
| **Mean rank** | 1.33 | 8.67 | 2.67 | 7 | 4 | 7.33 | 6.67 | 3 | 4 |
| **Ranking** | **1** | 9 | 2 | 7 | 4 | 8 | 6 | 3 | 4 |

Table 14 lists the three best indicators obtained by using each algorithm. In this section, we present a comparative analysis of the second half of the algorithms listed in Table 3 (physics- and human-based algorithms), highlighting the best results for each indicator in bold font. For the two-threshold segmentation problem, the nine algorithms have the same performance. This is because the two-dimensional search problem within the pixel interval [0–255] is relatively simple and the performance difference between the algorithms is not reflected. However, as the number of thresholds increases, the performance advantage of the proposed APO algorithm is presented. The experiment results illustrate that, in most cases, the proposed APO algorithm is superior to GSA, MVO, SCA, AOA, TLBO, SDO, SPO, and SPBO. According to the Friedman test, the overall ranking obtained by averaging the three indicators is as follows: APO ranks first, MVO ranks second, SPO ranks third, followed by AOA and SPBO with the same performance, SDO ranks sixth, SCA ranks seventh, TLBO ranks eighth, and GSA ranks ninth. Consequently, the proposed APO algorithm proves to be more effective than its counterparts on the multilevel image segmentation problem.

## 5. Conclusion

This study aimed to solve engineering optimization problems and provide flexible usage for practitioners. In this study, we proposed a new bio-inspired algorithm called artificial protozoa optimizer. It is inspired by natural protozoa, and the developed mathematical models mimic their survival behaviors, such as foraging, dormancy, and reproduction. It is worth mentioning that in the proposed APO algorithm, we introduced a mapping vector $M_f$ in the foraging behavior. This parameter determines the proportion of dimensional crossover of the candidate solutions. It is designed with the idea that "the better candidate solutions change less and the worse candidate solutions change more." This strategy, which favors exploitation for better candidates and exploration for worse candidates, can be extended to the design and improvement of metaheuristic algorithms. The effectiveness of the algorithm was validated through extensive experiments. Initially, the proposed APO was tested using the CEC2022 benchmark, which includes unimodal, multimodal, hybrid, and composition functions, enabling a comprehensive algorithm evaluation. Furthermore, considering a practical application context, the proposed APO was used to address five engineering design and multilevel image segmentation problems. Ultimately, the experimental data confirmed that the proposed algorithm significantly surpasses the compared state-of-the-art algorithms.

Despite the encouraging results, the algorithm introduced in this study is still at a preliminary stage when solving diverse and complex optimization problems. The following optimization problems and performance improvements can be considered in the future. First, the proposed APO algorithm can be explored to solve binary, large-scale, expensive, multi-objective, and multitask optimization problems. Second, several methods can be considered to enhance the performance of the algorithm, such as membrane computing, parallel algorithms, sampling techniques, clustering methods, and surrogate models. Finally, investigating hybridization using other algorithms is a promising research direction.

## CRediT authorship contribution statement

**Xiaopeng Wang:** Writing – original draft, Visualization, Software, Methodology, Investigation, Conceptualization. **Václav Snášel:** Writing – review & editing, Methodology, Investigation, Conceptualization. **Seyedali Mirjalili:** Visualization, Methodology, Writing – review & editing. **Jeng-Shyang Pan:** Conceptualization, Methodology, Writing – review & editing. **Lingping Kong:** Visualization, Software, Methodology, Data curation. **Hisham A. Shehadeh:** Software, Methodology, Data curation.

## Declaration of competing interest

The authors declare that they have no known competing financial interests or personal relationships that could have appeared to influence the work reported in this paper.

## Data availability

Data will be made available on request.

**Acknowledgments**

The authors gratefully acknowledge financial support CZ.02.01.01/ 00/22-008/0004590 by the Czech Republic Ministry of Education, Youth, and Sports in the project "Robotics and Advanced Industrial Production" (ROBOPROX).

**Appendix A. Supplementary data**

Supplementary material related to this article can be found online at https://doi.org/10.1016/j.knosys.2024.111737.